\pdfoutput=1

\documentclass[11pt]{article}

\usepackage{EMNLP2023}

\usepackage{times}
\usepackage{latexsym}

\usepackage{bm}
\usepackage{amsfonts}
\usepackage{amsmath}
\usepackage{mathrsfs}
\usepackage{graphicx}
\usepackage{booktabs}
\usepackage{multirow}
\usepackage{float}
\usepackage{subcaption}
\usepackage{stfloats}
\usepackage{amssymb}
\usepackage{xparse}
\usepackage{colortbl}
\usepackage{xspace}
\usepackage{enumitem}

\usepackage{soul} 
\usepackage{xcolor} 

\usepackage{array}

\usepackage[normalem]{ulem}
\useunder{\uline}{\ul}{}

\definecolor{lightblue}{RGB}{236,244,255}

\newcommand{\highlight}[1]{\sethlcolor{lightblue}\hl{#1}}

\usepackage[T1]{fontenc}
\DeclareFontShape{T1}{ptm}{m}{scit}{<->ssub * ptm/m/sc}{}

\usepackage{pifont}
\newcommand{\xmark}{\text{\ding{55}}}
\newcommand{\cmark}{\text{\ding{51}}}

\renewcommand*{\backref}[1]{}
\renewcommand*{\backrefalt}[4]{[{%
    \ifcase #1 Not cited.%
          \or Cited on page~#2.%
          \else Cited on pages #2.%
    \fi%
    }]}

\usepackage[normalem]{ulem}
\useunder{\uline}{\ul}{}

\newcommand{\mcal}[2][]{\mathcal{#2}\ifx\relax#1\relax\else^{\text{#1}}\fi}
\newcommand{\mbf}[2][]{\bm{\mathrm{#2}}\ifx\relax#1\relax\else^{\text{#1}}\fi}

\newcommand{\ls}{LibriSpeech\xspace}
\newcommand{\mustc}{\textsc{MuST-C v1.0}\xspace}
\newcommand{\mustcasr}{$\textsc{MuST-C}(\text{ASR})$\xspace}
\newcommand{\covost}{\textsc{CoVoST 2}\xspace}
\newcommand{\cv}{Common Voice\xspace}
\newcommand{\tstcommon}{{\fontfamily{qcr}\selectfont tst-COMMON}\xspace}
\newcommand{\test}{{\fontfamily{qcr}\selectfont test}\xspace}

\newcommand{\train}{{\fontfamily{qcr}\selectfont train}\xspace}
\newcommand{\wv}{\textsc{wav2vec 2.0}\xspace}
\newcommand{\zeroswot}{\textsc{ZeroSwot}\xspace}
\newcommand{\zeroswotm}{\textsc{ZeroSwot-Medium}\xspace}
\newcommand{\zeroswotl}{\textsc{ZeroSwot-Large}\xspace}
\newcommand{\med}{\textsc{Medium}\xspace}
\newcommand{\lrg}{\textsc{Large}\xspace}
\newcommand{\seamless}{\textsc{SeamlessM4T}\xspace}

\usepackage[T1]{fontenc}

\usepackage[utf8]{inputenc}

\usepackage{microtype}

\usepackage{inconsolata}

\title{Pushing the Limits of Zero-shot End-to-End Speech Translation}

\author{\phantom{-----------------} Ioannis Tsiamas, Gerard I. Gállego, José A. R. Fonollosa \\
    \phantom{-----------------} Universitat Politècnica de Catalunya, Barcelona \\
	\phantom{-----------------} \small\texttt{\{ioannis.tsiamas,gerard.ion.gallego,jose.fonollosa\}@upc.edu}
    \\\And \hspace{3.5cm} Marta R. Costa-jussà \\
    \hspace{3.5cm} FAIR Meta, Paris \\
	\hspace{3.5cm} \small\texttt{costajussa@meta.com}
 }

\begin{document}
\maketitle
\begin{abstract}

    Data scarcity and the modality gap between the speech and text modalities are two major obstacles of end-to-end Speech Translation (ST) systems, thus hindering their performance. Prior work has attempted to mitigate these challenges by leveraging external MT data and optimizing distance metrics that bring closer the speech-text representations. However, achieving competitive results typically requires some ST data. For this reason, we introduce \zeroswot, a method for zero-shot ST that bridges the modality gap without any paired ST data. Leveraging a novel CTC compression and Optimal Transport, we train a speech encoder using only ASR data, to align with the representation space of a massively multilingual MT model. The speech encoder seamlessly integrates with the MT model at inference, enabling direct translation from speech to text, across all languages supported by the MT model. Our experiments show that we can effectively close the modality gap without ST data, while our results on \textsc{MuST-C} and \textsc{CoVoST} demonstrate our method's superiority over not only previous zero-shot models, but also supervised ones, achieving state-of-the-art results.\footnote{\href{https://github.com/mt-upc/ZeroSwot}{https://github.com/mt-upc/ZeroSwot}}

\end{abstract}

\section{Introduction} \label{sec:intro}

    Traditionally, the standard approach for Speech translation (ST) has been the cascade model \cite{cascade_ney,cascade_mathias}, where Automatic Speech Recognition (ASR) and Machine Translation (MT) systems were chained together to produce the desired output. However, in recent years, there has been a paradigm shift towards end-to-end models, which aim to directly map the input speech to the target text without the intermediate transcription step \cite{e2e_berard}. Such models offer several advantages, including reduced error propagation, more compact architectures, and faster inference times \cite{taking_stock}.

    \begin{figure}
        \centering
        \includegraphics[width=0.4\textwidth]{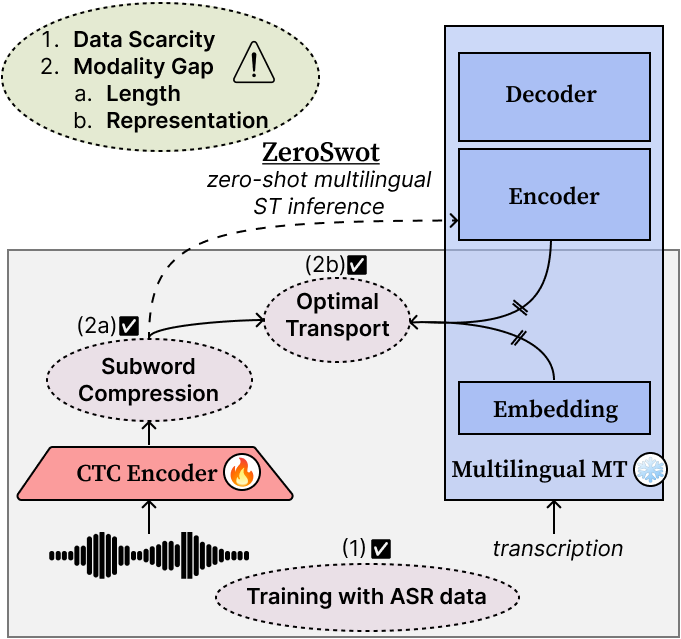}
        \caption{Proposed Zero-shot Speech Translation}
        \label{fig:introduction}
    \vspace{-0.5cm}
    \end{figure}

    Despite their advantages, end-to-end models require parallel ST data, which are often limited. To mitigate this \emph{data scarcity}, several strategies aim to leverage the more abundant ASR and MT data \cite{pretraining_berard,tied_multitask}. Then, another challenge that arises is the \emph{modality gap}, which emerges from the inherent differences between the speech and text modalities, manifesting in \emph{length mismatch} and \emph{representation mismatch}. Convolutional Length Adapters \cite{lna} or Connectionist Temporal Classification (CTC) \cite{ctc,bridging_the_modality_gap} can alleviate the length mismatch to some extent. To ease the representation gap, previous works have proposed a shared encoder \cite{xstnet} and optimizing distance metrics to bring the speech-text representation closer to the shared semantic space \cite{stemm,const}. Due to its ability to handle representations of different lengths, the Wasserstein distance \cite{wasserstein} using Optimal Transport, has emerged as an effective distance metric, specifically during pretraining \cite{ctc-ot,upc2023}. Although these methods show promise in reducing the modality gap and increasing translation quality, they still require at least some ST data.
    
    To this end, we propose \underline{Zero}-shot ST with \underline{S}ub\underline{w}ord compression and \underline{O}ptimal \underline{T}ransport (\zeroswot), a novel approach that aims to bridge the modality gap in a zero-shot fashion, only requiring ASR and MT data. Using Optimal Transport, we train a speech encoder based on \wv \cite{wav2vec2.0} to produce representations akin to the embedding space of a massively multilingual MT model, thus enabling zero-shot ST inference across all its supported target languages (Fig. \ref{fig:introduction}). To map the two embedding spaces, our method learns the tokenization of the MT model through CTC, and compresses the speech representations to the corresponding subwords, thus closing the length mismatch. Experiments on \textsc{MuST-C} \cite{mustc} reveal that our method is not only better than existing zero-shot models, by a large margin, but also surpasses supervised ones, achieving state-of-the-art results.  On CoVoST \cite{covost2}, \zeroswot outperforms the original version of the multimodal \textsc{SeamlessM4T} \cite{seamlessm4t}, while evaluations on the 88 target languages of FLEURS \cite{fleurs} showcase the massively multilingual capacity of our method. \zeroswot is also vastly superior to comparable CTC-based cascade ST models, and while it is on par with cascades that utilize strong attention-based encoder-decoder ASR models, it ranks better in terms of efficiency. Finally, we provide evidence of our method's ability to close the modality gap, both in terms of length and representation.
     
\section{Relevant Research} \label{sec:rel}

    \subsection{End-to-end Speech Translation} \label{subsec:rel.e2e_st}

        \setlength{\parindent}{0pt}

        \textbf{Data Scarcity.} To alleviate data scarcity, end-to-end ST methods usually utilize the more abundant data of ASR and MT. Such methods include pretraining \cite{pretraining_berard,pretraining_bansal,fairseq-s2t,pretraining_alinejad}, multi-task learning \cite{tied_multitask,task_aware}, data augmentation \cite{data_augmentation_jia,data_augmentation_pino,sample-translate-recombine,segaugment}, and knowledge distillation \cite{kd2019,kd2020}. Furthermore, many methods indirectly leverage external data through large foundation models \cite{foundation_models}. Initializing parts of the ST model with \wv \cite{wav2vec2.0} and \textsc{mBART50} \cite{mbart50} is a quite common practice \cite{lna,xstnet,chimera,upc2022}, while more recently \citet{naver_2023} utilized a massively multilingual MT model \cite{nllb} and \citet{lst} a large language model (LLaMa2) \cite{llama2}. Similarly, here employ \wv and \textsc{NLLB} \cite{nllb} to ease the data scarcity issue.

        
        \textbf{Modality Gap.} Implicit alignment techniques to bridge the representation gap involve a shared semantic encoder in a multitasking framework \cite{bridging_the_modality_gap,xstnet,general_framework}, which can be improved by mixup \cite{stemm,cmot}. In addition, explicit alignment methods optimize a distance metric to bring the speech-text representations closer in the shared semantic space. Such distance metrics include the euclidean \cite{bridging_the_modality_gap,joint_s2t}, cosine \cite{worse_wer_better_bleu}, and contrastive \cite{const,waco}, but they typically require some transformation in the representations, such as mean-pooling, while our approach optimizes a distance that does not alter the representation space. Methods to reduce the length discrepancy usually include sub-sampling the speech representation using convolutional length adaptors \cite{lna,upc2021,stemm,redapt} or character/phoneme-based CTC compression \cite{bridging_the_modality_gap,ctc_compression,sate}. Several methods have also used phonemized text in order to better match the representations in both length and content \cite{joint_s2t,stpt,ctc-ot}, but potentially limiting the quality of the text branch due to noise. In this work, we introduce a novel CTC-based compression to subword units, directly aligning with the tokenization of MT models.

        \setlength{\parindent}{12pt}

    \subsection{Optimal Transport} \label{subsec:rel.ot}

        Optimal Transport (OT) \cite{ot}, traditionally used in NLP and MT \cite{ot_chen,ot_alqahtani}, has recently also been applied to ST. \citet{cmot} used OT to find the alignment between speech and text features to apply Mixup. \citet{ctc-ot} proposed to use OT in a siamese pretraining setting in combination with CTC, yielding improvements compared to the standard ASR pretraining, while also proposing a positional regularization in order to make OT applicable to sequences. \citet{upc2023} extended this pretraining in the context of foundation models, while also freezing the text branch and using CTC compression in order to achieve better adaptation with the text decoder during ST finetuning. Inspired by these works, we also follow the siamese paradigm, but as a means of directly integrating the speech encoder to the text space of an MT model, thus not requiring any ST data.

    \subsection{Zero-shot Speech Translation} \label{subsec:rel.zs_st}

    \citet{language_specific_speech} were the first to achieve zero-shot ST, by incorporating a speech encoder to the representation space of an MT model using language-specific encoders-decoders \cite{language_specific_mt}. \citet{tackling_data_scarcity} combined multi-task learning on ASR and MT, introducing an auxiliary loss to minimize the euclidean distance between mean-pooled speech and text representations. T-modules \cite{t_modules} harnessed mined data to learn a fixed-size multilingual and multimodal representation space, subsequently enabling zero-shot ST encoding and decoding from that space. DCMA \cite{dcma} projects both speech and text into a fixed-length joint space using an attention-driven shared memory module, after which it enforces the speech distribution of a codebook to closely mirror its textual counterpart. \citet{simreg} proposed a strategy that employs a multimodal variant of cross-lingual consistency regularization \cite{cross_con_st}, training a model with \wv and a shared encoder on both MT and ASR data. Recently, \citet{towards-zero} introduced a zero-shot ST training method by adapting a new speech encoder to bilingual MT models using OT and a CTC module that shares the vocabulary of the MT model, attaining improvements over zero-shot methods and CTC-based cascade systems. In contrast, our approach focuses on achieving multilingual zero-shot ST, by adapting a large CTC-based speech encoder \cite{wav2vec2.0} to a massively multilingual MT model \cite{nllb}, aiming for high-quality results comparable to those of state-of-the-art cascade and end-to-end models. To address the unique challenges associated with adapting to the large vocabulary size of multilingual models (e.g., 250k tokens for \cite{nllb}), we propose a novel compression method that effectively learns the subword tokenization of the MT model, rather than predicting actual tokens. This decoupling from the vocabulary size ensures efficiency and scalability, overcoming limitations that may arise in previous methods \cite{towards-zero}, due to the need of sharing the MT vocabulary with the CTC module.
    
\section{Methodology} \label{sec:meth}

    \begin{figure*}[ht]
        \centering
        \includegraphics[width=\textwidth]{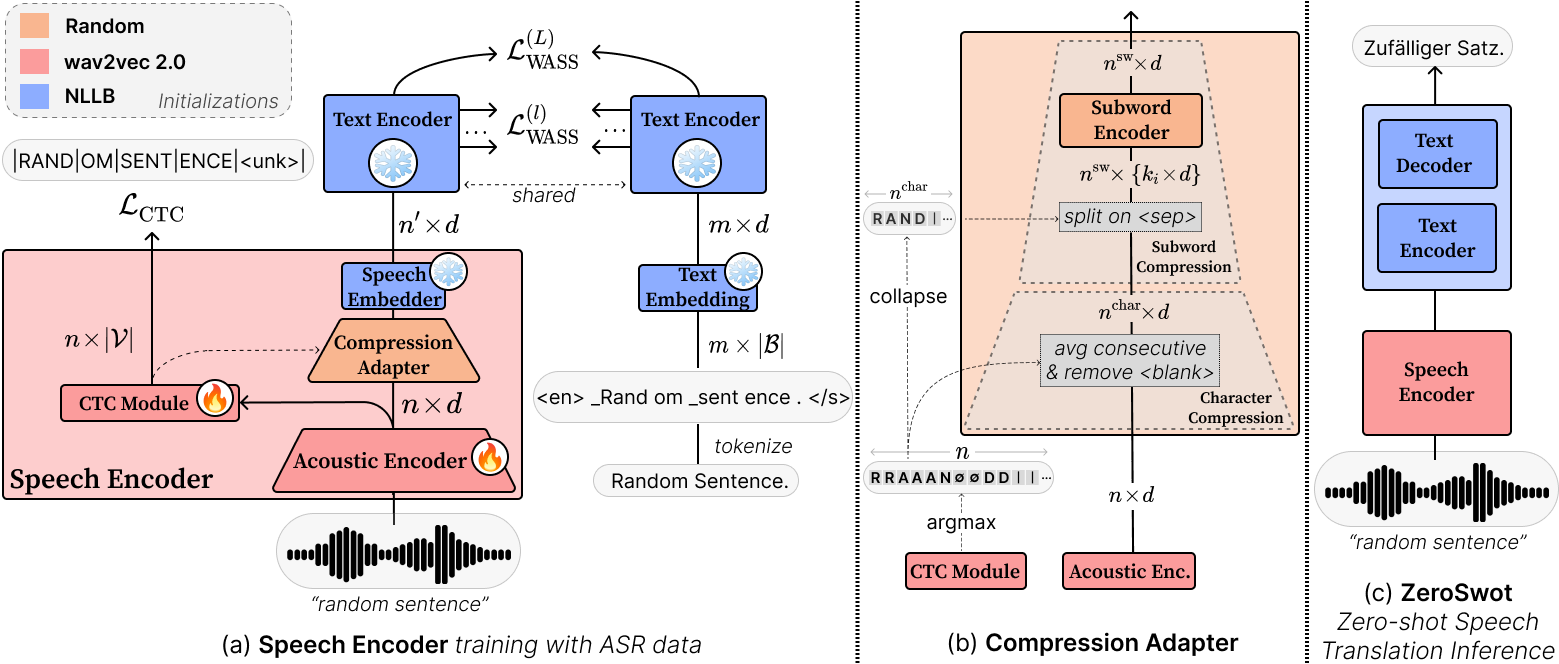}
        \caption{Methodology: Speech Encoder Training, Compression Adapter, and Inference with \zeroswot.}
        \label{fig:methodology}
    \end{figure*}

    \subsection{Problem Definition \& Proposed Solution}

        \setlength{\parindent}{0pt}

        \textbf{End-to-end Speech Translation.} A speech translation corpus consists of speech-transcription-translation triplets, $\mcal{D}=\{(\mbf{s}, \mbf{x}, \mbf{y})\}$, where $\mbf{s}$ is the speech signal, $\mbf{x}$ is the transcription, and $\mbf{y}$ the translation. End-to-end ST systems aim to learn a model $\mcal[ST]{M}$ that generates the translation $\mbf{y}$ directly from the speech $\mbf{s}$, using the parallel ST data $\mcal[ST]{D}=\{(\mbf{s}, \mbf{y})\}$. The transcription $\mbf{x}$ is optionally utilized, as in pretraining or multi-task learning.
        
        \textbf{Zero-shot Speech Translation.} In a zero-shot ST setting, we aim to learn a model $\mcal[ZS-ST]{M}$ that generates the translation $\mbf{y}$ with access to corpora $\mcal[MT]{D} = \{(\mbf{x}, \mbf{y})\}$ and $\mcal[ASR]{D} = \{(\mbf{s}, \mbf{x})\}$, but without directly utilizing the parallel corpus $\mcal[ST]{D}$.

        \textbf{Proposed Solution.} Assuming access to an MT model $\mcal[MT]{M}$ obtained via $\mcal[MT]{D}$, we utilize $\mcal[ASR]{D}$ to learn a speech encoder $\mbf{S\mbox{-}Enc}$ that produces representations akin to those expected by $\mcal[MT]{M}$. On inference time, we achieve zero-shot ST by replacing the embedding layer of $\mcal[MT]{M}$ with $\mbf{S\mbox{-}Enc}$.

        \setlength{\parindent}{12pt}

    \subsection{Model Architecture} \label{subsec:meth.spch_enc}

        Our model architecture includes a speech and a text branch. The text branch consists of a text encoder, which remains frozen during training. The speech branch consists of an acoustic encoder, a CTC module, a compression adapter, a speech embedder and a semantic encoder, of which the parameters are shared with the frozen text encoder (Figure \ref{fig:methodology}a).

        \subsubsection{Text Encoder} \label{subsec:meth.mt}

            The text encoder is a Transformer, of which the parameters are initialized from a massively multilingual MT model \cite{nllb}, and kept frozen during training. The tokenized\footnote{With prepended \texttt{<lang>} and appended \texttt{</s>} tokens.} source text $\mbf{x}$ with length $m$ is passed through an embedding layer $\mbf{Emb}\!:\!|\mathcal{B}| \rightarrow d$ to obtain $\mbf[x]{e} \in \mathbb{R}^{m \times d}$, where $\mathcal{B}$ is a multilingual subword-level vocabulary. Sinusoidal positional encodings $\mbf{Pos}(\cdot)$ are added after an up-scaling \cite{transformer}, and a Transformer encoder $\mbf{T\mbox{-}Enc}$ with $L$ layers is used to obtain a semantic representation $\mbf[x]{h}_L \in \mathbb{R}^{m \times d}$.
            \begin{align}
                \mbf[x]{e} &= \sqrt{d} \cdot \mbf{Emb}(\mbf{x}) + \mbf{Pos}(m) \label{eq:text_emb} \\
                \mbf[x]{h}_L &= \mbf{T\mbox{-}Enc}(\mbf[x]{e}) \label{eq:text_enc}
            \end{align}

        \subsubsection{Acoustic Encoder} \label{subsec:meth.acoustic}

            The acoustic encoder $\mbf{A\mbox{-}Enc}$ takes as input the speech signal $\mbf{s} \in \mathbb{R}^{\ell}$ and produces an acoustic representation $\mbf{a} \in \mathbb{R}^{n \times d}$. It consists of a series of strided convolutional layers that downsample the signal by $r$ (thus, $n\!=\!\ell/r)$, followed by a Transformer encoder with $N$ layers. It is initialized from \wv and is finetuned during training.
            \begin{align}
                \mbf{a} &= \mbf{A\mbox{-}Enc}(\mbf{s}) \label{eq:acoustic_encoder}
            \end{align}

        \subsubsection{Connectionist Temporal Classification}  \label{subsec:meth.ctc}

            We use a CTC loss \cite{ctc} to maintain a meaningful acoustic representation and provide the necessary information to the compression adapter (\S \ref{subsec:meth.compress}). A CTC module, which is a linear softmax layer produces probabilities $\mbf{p} \in \mathbb{R}^{n \times |\mathcal{V}|}$, where $\mathcal{V}$ is a letter-based vocabulary.\footnote{Incl.
            \emph{blank} \texttt{<blank>}, \emph{unknown} \texttt{<unk>}, \emph{separator} \texttt{<sep>}.}
            \begin{align}
                \mbf{p} &= \text{softmax}(\mbf{a} \cdot \mbf[ctc]{W}  + \mbf[ctc]{b})
            \end{align}
            
            Where $\mbf[ctc]{W} \in \mathbb{R}^{d \times |\mathcal{V}|}$ and $\mbf[ctc]{b} \in \mathbb{R}^{|\mathcal{V}|}$ are trainable parameters, initialized from \wv.

            Given the CTC labels $\mbf{z}$, and their conditional probability $P(\mbf{z}|\mbf{p})$, which is obtained by summing over the probability of all possible alignment paths between the acoustic representation $\mbf{a}$ and $\mbf{z}$, the CTC loss is defined as:
            \begin{align}
                \mathcal{L}_\text{CTC} &= - \log P(\mbf{z}|\mbf{p}) \label{eq:ctc}
            \end{align}

            \begin{table}[ht]
                \centering
                \resizebox{\columnwidth}{!}{
                \begin{tabular}{@{}ccc@{}}
                \toprule
                \textbf{Separation} & \textbf{Unk} & \textbf{CTC labels for} \emph{"Random Sentence."}                         \\ \midrule
                words          & $\xmark$             & R A N D O M $\boldsymbol{|}$ S E N T E N C E $\boldsymbol{|}$                                          \\
                subwords       & $\cmark$             & R A N D $\boldsymbol{|}$ O M $\boldsymbol{|}$ S E N T $\boldsymbol{|}$ E N C E $\boldsymbol{|}$ \texttt{<unk>} $\boldsymbol{|}$                              \\ \bottomrule
                \end{tabular}
                }
                \caption{Standard vs proposed CTC labels example.}
                \label{tab:ctc_labels}
            \end{table}

            The labels $\mbf{z}$ are usually obtained from the transcription $\mbf{x}$ via splitting on words and removing characters absent from $\mathcal{V}$, such as punctuation. But this implies a word-level tokenization and text normalization, which is inconsistent with the text branch (\S \ref{subsec:meth.mt}). Thus, in order to minimize the discrepancies between the two branches, we propose to split on subwords using the text branch tokenizer, and keep the positions of characters not in $\mathcal{V}$, by replacing them with the unknown token (Table \ref{tab:ctc_labels}). This is particularly important for the effectiveness of our compression to subwords (\S \ref{subsec:meth.compress}).

        \subsubsection{Compression Adapter}  \label{subsec:meth.compress}
        
            \begin{figure}[ht]
                \centering
                \includegraphics[width=\columnwidth]{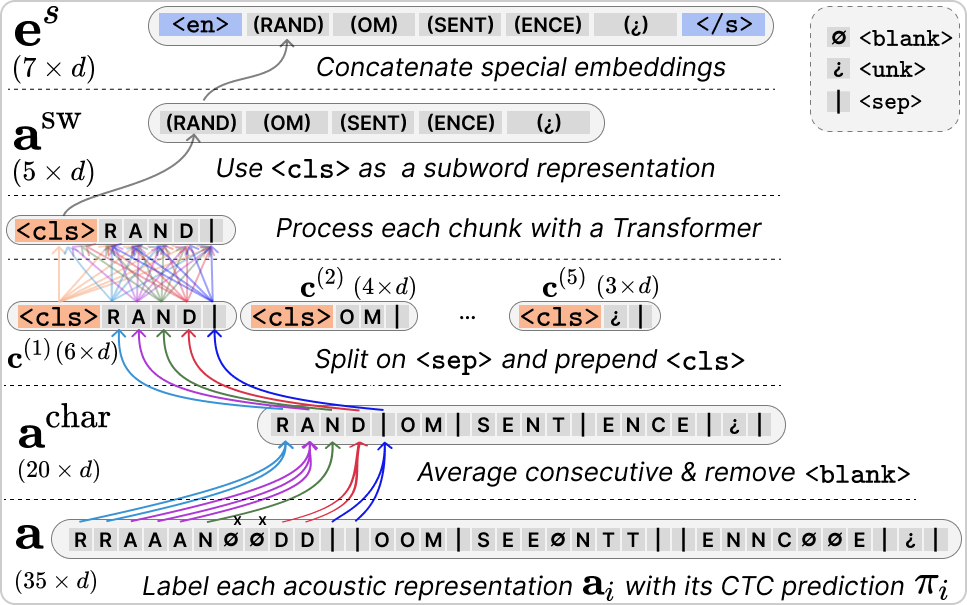}
                \caption{Example of the proposed compression.}
                \label{fig:compression_example}
            \end{figure}

            The compression adapter (Fig. \ref{fig:methodology}b) takes as input the acoustic representation $\mbf{a}$ and the CTC probabilities $\mbf{p}$, to produce a compressed acoustic representation $\mbf[sw]{a} \in \mathbb{R}^{n^\text{sw} \times d}$ (where $n^\text{sw} \! \leq \! n$) that resembles in length and content the text embedding $\mbf[x]{e}$ (Eq. \ref{eq:text_emb}). It involves two stages: compression to characters \cite{bridging_the_modality_gap}, followed by a novel compression to subwords.

            \setlength{\parindent}{0pt}
            
            \textbf{Character Compression.} Each vector $\mbf{a}_i \in \mathbb{R}^d$ of the acoustic representation is labeled according to the corresponding greedy CTC prediction $\pi_i \! = \! \text{argmax}(\mbf{p}_i)$. Then, consecutive same-labeled representations are grouped together, and combined via mean-pooling, while removing those labeled as blanks, thus having a character-like representation $\mbf[char]{a} \! \in \! \mathbb{R}^{n^\text{char} \times d}$, where $n^\text{char} \! \leq \! n$. Similarly, $\mbf{p}$ is compressed to $\mbf[char]{p} \in \mathbb{R}^{n^\text{char} \times |\mathcal{V}|}$.
            \begin{align}
                \mbf{\pi} &= \text{argmax}(\mbf{p}) \\
                \mbf[char]{p} &= \texttt{char-compress}(\mbf{p}| \mbf{\pi}) \\
                \mbf[char]{a} &= \texttt{char-compress}(\mbf{a}| \mbf{\pi})
            \end{align}

            \textbf{Subword Compression.} The character representation $\mbf[char]{a}$ is split into chunks $\mbf{c}^{(1)}, \dots, \mbf{c}^{(n^\text{sw})}$ according to the positions labeled as separators, where $n^\text{sw}$ is the number of predicted separators, $\mbf{c}^{(i)} \in \mathbb{R}^{k_i \times d}$ is the $i-$th chunk representation (with $k_i$ number of characters), and $n^\text{char} \! = \! \sum_{i=0}^{n^\text{sw}} k_i$. Since the CTC is predicting the tokenization of the text branch through the proposed target labels, each chunk is combining information that resemble subword tokens from $\mathcal{V}$. Each chunk of character representations is processed independently with $\mbf{Subword\mbox{-}Enc}: \mathbb{R}^{k_i \times d} \rightarrow \mathbb{R}^d$, which is Transformer encoder that prepends a trainable \texttt{<cls>} token to each chunk, which is then used a subword representation of the whole chunk \cite{bert}. Then, by concatenation we obtain a subword-like compressed representation $\mbf[sw]{a} \in \mathbb{R}^{n^\text{sw} \times d}$, where $n^\text{sw} \! \leq \! n^\text{char} \! \leq \! n$ (Fig. \ref{fig:compression_example}).
           \begin{align}
                 \mbf{c}^{(1)}&, \dots, \mbf{c}^{(n^\text{sw})} = \texttt{split}(\mbf[char]{a}| \mbf[char]{\pi}) \\
                 \mbf[sw]{a}_i &= \mbf{Subword\mbox{-}Enc}\bigl(\mbf{c}^{(i)}\bigr)
            \end{align}
            
            \setlength{\parindent}{12pt}

        \subsubsection{Speech Embedder} \label{sec:meth.speech_embedder}

            We concatenate to the compressed acoustic representation two vectors $\mbf[\texttt{<lang>}]{\epsilon}, \mbf[\texttt{</s>}]{\epsilon} \! \in \! \mathbb{R}^d$ that function as the source language and end of sentence embeddings. Similar to the text branch, we scale the representation by $\sqrt{d}$ and add sinusoidal positional encodings with $\mbf{Pos}(\cdot)$ (Eq. \ref{eq:text_emb}), thus obtaining a speech embedding $\mbf[s]{e} \in \mathbb{R}^{n' \times d}$, where $n' \! = \! n^\text{sw} + 2$. The special embeddings are initialized from the text embedding (Eq. \ref{eq:text_emb}), and remain frozen.

        \subsubsection{Semantic Encoder} \label{sec:meth.semantic}

            The speech embedding $\mbf[s]{e}$ is passed through a Transformer encoder $\mbf{T\mbox{-}Enc}$, which is frozen and shared with the text branch (Eq. \ref{eq:text_enc}), to obtain a semantic speech representation $\mbf[s]{h}_L \in \mathbb{R}^{n' \times d}$.
            \begin{align}
                \mbf[s]{e} &= \sqrt{d} [\mbf[\texttt{<lang>}]{\epsilon}, \mbf[sw]{a}, \mbf[\texttt{<s>}]{\epsilon}] + \mbf{Pos}(n') \\
                \mbf[s]{h}_L &= \mbf{T\mbox{-}Enc}(\mbf[s]{e})
            \end{align}
    
    \subsection{Optimal Transport} \label{subsec:meth.ot}

        To align a speech representation $\mbf[s]{h} \in \mathbb{R}^{n' \times d}$ to the text representation $\mbf[x]{h} \in \mathbb{R}^{m \times d}$, we are minimizing their Wasserstein loss \cite{wasserstein} using Optimal Transport (OT) \cite{ot}, as in \citet{ctc-ot,upc2023}.

        We assume two uniform probability distributions $\mbf[s]{\phi}, \mbf[x]{\phi}$, with $\mbf[s]{\phi}_i \! = \! 1/n'$ and $\mbf[x]{\phi}_j \! = \! 1/m$, that define the mass of each position in the speech and text representations. Using a squared euclidean cost function we obtain a pairwise cost matrix $\mbf{C} \in \mathbb{R}_+^{n' \times m}$, where $\mbf{C}_{ij} \! = \! ||\mbf[s]{h}_i - \mbf[x]{h}_j||_2^2$. Then, the Wasserstein distance $W_\delta \in \mathbb{R}_+$ is defined as the minimum transportation cost over all possible transportation plans $\mbf{Z} \in \mathbb{R}_+^{n' \times m}$ of the total masses $\mbf[s]{\phi}, \mbf[x]{\phi}$. The optimization is thus subjected to constraints $\sum_{j=1}^m \mbf{Z}_{:,j} \! = \! 1/n'$ and $\sum_{i=1}^{n'} \mbf{Z}_{i,:} \! = \! 1/m$.
        \begin{equation}
            W_\delta = \underset{\mbf{Z}}{\text{min}} \sum_{i=1}^{n'} \sum_{j=1}^m \mbf{Z}_{ij} \mbf{C}_{ij}
        \end{equation}

         In practice, to make the Wasserstein distance differentiable and efficient to compute using the Sinkhorn algorithm \cite{sinkhorn}, we use its entropy-regularized upper-bound estimation. Furthermore, since the Wasserstein distance is permutation invariant, we follow \citet{ctc-ot} and incorporate two positional regularization vectors $\mbf[s]{v} \in \mathbb{R}^{n'}$ and $\mbf[x]{v} \in \mathbb{R}^{m}$ in $\mbf[s]{h}$ and $\mbf[x]{h}$ accordingly, thus penalizing the transportation of mass between two distant positions $i$ and $j$.
        \begin{align}
            \mbf[s]{\bar{h}} &= [\mbf[s]{h}; \mu \mbf[s]{v}] \text{\phantom{a}, where 
            \hphantom{a}} \mbf[s]{v}_i = \frac{i-1}{n'-1} \label{eq:modified_repr_s} \\
            \mbf[x]{\bar{h}} &= [\mbf[x]{h}; \mu \mbf[x]{v}] \text{\phantom{a}, where 
            \hphantom{a}} \mbf[x]{v}_j = \frac{j-1}{m-1} \label{eq:modified_repr_x}
        \end{align}

        And thus for the positional regularized cost matrix $\mbf{\bar{C}}$, the upper-bound Wasserstein distance that we use as a loss function is defined as:
        \begin{equation} \label{eq:wass}
            \mathcal{L}_\text{WASS} = \underset{\mbf{Z}}{\text{min}} \Bigl( \sum_{i=1}^{n'} \sum_{j=1}^m \mbf{Z}_{ij} \mbf{\bar{C}}_{ij} - \lambda \text{H}(\mbf{Z}) \Bigr)
        \end{equation}

        Where $\lambda > 0$ is a hyperparameter, and $\text{H}(\cdot)$ denotes the von Neumann entropy of a matrix.

    \subsection{Final Loss} \label{subsec:meth.final_loss}

        We apply a Wasserstein loss (Eq. \ref{eq:wass}), not only at the final layer but also at intermediate ones, since they also carry important information, and can thus help producing more robust speech representations. Thus, the final loss is defined as:
        \begin{align}
            \mathcal{L} &= \sum_{l \in \mathcal{I}} \frac{\alpha}{|\mathcal{I}|} \mathcal{L}_\text{WASS}^{(l)} + (1-\alpha)\mathcal{L}_\text{CTC} \label{eq:final_loss}
        \end{align}

        Where $\mathcal{I}$ is the set of shared encoder layers for which we apply Wasserstein losses, and $\alpha > 0$ is a hyperparameter to balance the relative importance between the Wasserstein and CTC losses.
        
    \subsection{Zero-shot Speech Translation Inference}

        To enable zero-shot ST inference, we simply replace the MT embedding layer with the trained speech encoder. (\zeroswot, Fig. \ref{fig:methodology}c).

\section{Experimental Setup} \label{sec:exp}
    \setlength{\parindent}{0pt}

    \textbf{Data.} For training we are using \cv \cite{commonvoice}, the speech-transcription pairs of \mustc \cite{mustc}, and \ls \cite{librispeech}. We evaluate our models on three multilingual ST benchmarks: \mustc (En $\rightarrow$ 8 lang.), \covost \cite{covost2} (En $\rightarrow$ 15 lang.), and FLEURS \cite{fleurs} (En $\rightarrow$ 88 lang.) (Appendix \ref{appendix:data}).

    \textbf{Model Architecture.} The acoustic encoder follows \wv \lrg \cite{wav2vec2.0} with $N\!=\!24$ layers and dimensionality $d\!=\!1024$ (315M parameters). The CTC module uses a vocabulary with size $|\mathcal{V}|\!=\!30$, and together with the acoustic encoder are initialized from a CTC-finetuned version of \wv on LibriSpeech \cite{librispeech}. The subword encoder has 3 layers with dimensionality of 1024 (35M). The text encoder has either $L\!=\!12$ or $L\!=\!24$, both with dimensionality 1024, and an embedding size of $|\mathcal{B}|\!=\! 256k$. We initialize its parameters from two dense versions of NLLB \cite{nllb}, the 600M (\med) or the 1.3B (\lrg). Based on the size of NLLB, the resulting \zeroswot models used for inference have 0.95B (\med) or 1.65B (\lrg) parameters, while both of them the have the same number of training parameters (350M). We also train a \textsc{Small} version based on \wv \textsc{Base} and NLLB-\med (Appendix \ref{appendix:zeroswot_architecture}).

    \textbf{Training details.} We use AdamW \cite{adamw} with a learning rate of 3e-4, inverse square root scheduler with warmup, and batch size of 32M tokens. We apply dropout of 0.1 as well as time and channel masking \cite{wav2vec2.0}. Speech input is 16kHz waveforms, and text is tokenized with sentencepiece \cite{sentencepiece}, using the NLLB model (Appendix \ref{appendix:zeroswot_training}).

    \textbf{Loss.} We set $\alpha \! = \! 0.9$ and apply Wasserstein losses for $\mathcal{I}\!=\!\{6,7,8,9,10,11,12\}$ (\med) and $\mathcal{I}\!=\!\{12,14,16,18,20,22,24\}$ (\lrg) (Eq. \ref{eq:final_loss}). Positional and entropy regularization are set to $\mu\!=\!10$ (Eq. \ref{eq:modified_repr_s}, \ref{eq:modified_repr_x}), and $\lambda\!=\!1$ (Eq. \ref{eq:wass}).

    \textbf{Evaluation.} We apply checkpoint averaging and use a beam search of size 5. We evaluate with case-sensitive detokenized BLEU \cite{bleu} using sacreBLEU \cite{sacrebleu} (Appendix \ref{appendix:evaluation}).

    \setlength{\parindent}{12pt}

\section{Results} \label{sec:res}

    \setlength{\parindent}{0pt}

        \textbf{\mustc{}.} In Table \ref{tab:main_results_mustc} we present the results of our proposed \zeroswot in \textsc{MuST-C} and compare them with other SOTA zero-shot and supervised methods. We train \zeroswot under three scenarios, regarding data usage (Table \ref{tab:data_usage}), and in three different sizes depending on the \wv and NLLB versions (Table \ref{tab:model_size}). For scenario C we first multilingually finetuned NLLB on \textsc{MuST-C} En-X MT data (Appendix \ref{appendix:mt}), before adapting to it the speech encoder with \textsc{MuST-C} ASR data. 

        \begin{table}[H]
        \centering
            \resizebox{\columnwidth}{!}{
            \begin{tabular}{@{}cccc@{}}
            \toprule
            \textbf{Scenario} & \textbf{MuST-C Data}                   & \textbf{ASR} & \textbf{NLLB} \\ \midrule
            A                      & $\text{ASR} \xmark / \text{MT} \xmark$ & CommonVoice  & Original              \\
            B                      & $\text{ASR} \cmark / \text{MT} \xmark$ & \textsc{MuST-C}       & Original              \\
            C                      & $\text{ASR} \cmark / \text{MT} \cmark$ & \textsc{MuST-C}       & Mult. FT         \\ \bottomrule
            \end{tabular}
            }
            \caption{Data usage scenarios in \textsc{MuST-C.}}
            \label{tab:data_usage}
        \end{table}
        \vspace{-0.3cm}
        \begin{table}[H]
        \centering
            \resizebox{\columnwidth}{!}{
                \begin{tabular}{@{}cccc@{}}
                \toprule
                \textbf{\begin{tabular}[c]{@{}c@{}}\textsc{ZeroSwot}\\ Model Size\end{tabular}} & \textbf{\begin{tabular}[c]{@{}c@{}}Train/Infer\\ Parameters\end{tabular}} & \textbf{\wv}   & \textbf{NLLB} \\ \midrule
                \textsc{Small}                                                                  & 0.1B / 0.7B                                                               & \textsc{Base} (90M)  & 600M          \\
                \textsc{Medium}                                                                 & 0.35B / 0.95B                                                             & \textsc{Large} (300M) & 600M          \\
                \textsc{Large}                                                                  & 0.35B / 1.65B                                                             & \textsc{Large} (300M) & 1.3B          \\ \bottomrule
                \end{tabular}
            }
            \caption{Model sizes for \zeroswot.}
            \label{tab:model_size}
        \end{table}

        In the upper part of Table \ref{tab:main_results_mustc} we compare our small \textsc{ZeroSwot} model using only ASR data from \textsc{MuST-C} (scenario B) to previously proposed zero-shot methods. Our results show that our model can surpass all previous methods, even though some of them use both ASR and MT data from MuST-C, and also have more training parameters. Then, comparing our larger versions to supervised methods, \zeroswot sets new SOTA in 7/8 directions, with multiple versions of our model outperforming the previous best. Notably, \zeroswotl(C) is better in En-De and En-Es than the recently proposed LST \cite{lst}, which is based on LLaMA2-13B \cite{llama2}, and compared to our method has $40\!\times/\text{ }8\times$ more training/inference parameters. 


        \begin{table*}[ht]
            \centering
            \resizebox{\textwidth}{!}{
            \begin{tabular}{@{}lrcccccccccccc@{}}
            \toprule \toprule
            \multicolumn{2}{c}{}                                                              & \multicolumn{2}{c}{\textbf{MuST-C Data}} &                                                                                          & \multicolumn{9}{c}{\textbf{BLEU}}                                                                                                                                                                                                                                                                                                                                                          \\ \cmidrule(lr){3-4} \cmidrule(l){6-14} 
            \multicolumn{2}{c}{\multirow{-2}{*}{\textbf{Models}}}                             & \textbf{ASR}        & \textbf{MT}        & \multirow{-2}{*}{\textbf{\begin{tabular}[c]{@{}c@{}}Train/Infer\\ Size (B)\end{tabular}}} & \textbf{De}                           & \textbf{Es}                           & \textbf{Fr}                           & \textbf{It}                           & \textbf{Nl}                           & \textbf{Pt}                           & \textbf{Ro}                           & \textbf{Ru}                                                & \textbf{Average}                      \\ \midrule
            \multicolumn{14}{c}{\textit{Previous Zero-shot End-to-End ST compared to our small model}}                                                                                                                                                                                                                                                                                                                                                                                                                                                                                               \\
            \multicolumn{2}{l|}{T-Modules \cite{t_modules}}                                   & $\cmark$            & $\xmark$           & \multicolumn{1}{c|}{$2^*$}                                                               & 23.8                                  & 27.4                                  & 32.7                                  & -                                     & -                                     & -                                     & -                                     & \multicolumn{1}{c|}{-}                                     & -                                     \\
            \multicolumn{2}{l|}{DCMA \cite{dcma}}                                             & $\cmark$            & $\xmark$           & \multicolumn{1}{c|}{$0.15^*$}                                                            & 22.4                                  & 24.6                                  & 29.7                                  & 18.4                                  & 22.0                                  & 24.2                                  & 16.8                                  & \multicolumn{1}{c|}{11.8}                                  & 21.2                                  \\
            \multicolumn{2}{l|}{DCMA \cite{dcma}}                                             & $\cmark$            & $\cmark$           & \multicolumn{1}{c|}{$0.15^*$}                                                            & 24.0                                  & 26.2                                  & 33.1                                  & 24.1                                  & 28.3                                  & 29.2                                  & 22.2                                  & \multicolumn{1}{c|}{16.0}                                  & 25.4                                  \\
            \multicolumn{2}{l|}{SimZeroCR \cite{simreg}}                                      & $\cmark$            & $\cmark$           & \multicolumn{1}{c|}{$0.15^*$}                                                            & 25.1                                  & 27.0                                  & 34.6                                  & -                                     & -                                     & -                                     & -                                     & \multicolumn{1}{c|}{15.6}                                  & -        \\
            \multicolumn{2}{l|}{Towards-Zero-shot \cite{towards-zero}}                                      & $\cmark$            & $\xmark$           & \multicolumn{1}{c|}{$0.1$}                                                            & 23.4                                  & 26.5                                  & 33.7                                  & -                                     & -                                     & -                                     & -                                     & \multicolumn{1}{c|}{-}                                 & -        \\
            \multicolumn{2}{l|}{Towards-Zero-shot \cite{towards-zero}}                                      & $\cmark$            & $\cmark$           & \multicolumn{1}{c|}{$0.1$}                                                            & 26.5                                  & 29.5                                  & 35.3                                  & -                                     & -                                     & -                                     & -                                     & \multicolumn{1}{c|}{-}                                 & -      
            \\ \midrule
            \phantom{ai}\textsc{ZeroSwot-Small}                    & \multicolumn{1}{r|}{(B)} & $\cmark$            & $\xmark$           & \multicolumn{1}{c|}{$0.1 / 0.7$}                                           & \textbf{27.3}                         & \textbf{31.7}                         & \textbf{35.8}                         & \textbf{26.8}                         & \textbf{30.9}                         & \textbf{31.6}                         & \textbf{25.3}                         & \multicolumn{1}{c|}{\textbf{17.8}}                         & \textbf{28.4}                         \\ \midrule \midrule \\ \midrule \midrule
            \multicolumn{14}{c}{\textit{Supervised End-to-End ST compared to our larger Zero-shot ST models}}                                                                                                                                                                                                                                                                                                                                                                                                                                                                                                                             \\
            \multicolumn{2}{l|}{Chimera \cite{chimera}}                                       & \multicolumn{2}{c}{$\cmark$}             & \multicolumn{1}{c|}{$0.15^{\phantom{*}}$}                                                & 27.1                                  & 30.6                                  & 35.6                                  & 25.0                                  & 29.2                                  & 30.2                                  & 24.0                                  & \multicolumn{1}{c|}{17.4}                                  & 27.4                                  \\
            \multicolumn{2}{l|}{STEMM \cite{stemm}}                                           & \multicolumn{2}{c}{$\cmark$}             & \multicolumn{1}{c|}{$0.15^*$}                                                            & 28.7                                  & 31.0                                  & 37.4                                  & 25.8                                  & 30.5                                  & 31.7                                  & 24.5                                  & \multicolumn{1}{c|}{17.8}                                  & 28.4                                  \\
            \multicolumn{2}{l|}{SpeechUT \cite{speechut}}                                     & \multicolumn{2}{c}{$\cmark$}             & \multicolumn{1}{c|}{$0.15^{\phantom{*}}$}                                                & 30.1                                  & 33.6                                  & 41.4                                  & -                                     & -                                     & -                                     & -                                     & \multicolumn{1}{c|}{-}                                     & -                                     \\
            \multicolumn{2}{l|}{Siamese-PT \cite{ctc-ot}}                                     & \multicolumn{2}{c}{$\cmark$}             & \multicolumn{1}{c|}{$0.25^*$}                                                            & 27.9                                  & 31.8                                  & 39.2                                  & 27.7                                  & 31.7                                  & {\ul 34.2}                            & {\ul 27.0}                            & \multicolumn{1}{c|}{18.5}                                  & 29.8                                  \\
            \multicolumn{2}{l|}{CRESS \cite{cress}}                                           & \multicolumn{2}{c}{$\cmark$}             & \multicolumn{1}{c|}{$0.15^*$}                                                            & 29.4                                  & 33.2                                  & 40.1                                  & 27.6                                  & 32.2                                  & 33.6                                  & 26.4                                  & \multicolumn{1}{c|}{19.7}                                  & 30.3                                  \\
            \multicolumn{2}{l|}{SimRegCR \cite{simreg}}                                       & \multicolumn{2}{c}{$\cmark$}             & \multicolumn{1}{c|}{$0.15^*$}                                                            & 29.2                                  & 33.0                                  & 40.0                                  & {\ul 28.2}                            & {\ul 32.7}                            & {\ul 34.2}                            & 26.7                                  & \multicolumn{1}{c|}{{\ul 20.1}}                            & {\ul 30.5}                            \\
            \multicolumn{2}{l|}{LST (LLaMA2-13B) \cite{lst}}                                  & \multicolumn{2}{c}{$\cmark$}             & \multicolumn{1}{c|}{$13$}                                                                  & {\ul 30.4}                            & {\ul 35.3}                            & {\ul \textbf{41.6}}                   & -                                     & -                                     & -                                     & -                                     & \multicolumn{1}{c|}{-}                                     & -                                     \\ \midrule
                                                                   & \multicolumn{1}{r|}{(A)} & $\xmark$            & $\xmark$           & \multicolumn{1}{c|}{$0.35 / 0.95^{\phantom{\dagger}}$}                                          & 24.8                                  & 30.0                                  & 32.6                                  & 24.1                                  & 28.6                                  & 28.8                                  & 22.9                                  & \multicolumn{1}{c|}{16.4}                                  & 26.0                                  \\
                                                                   & \multicolumn{1}{r|}{(B)} & $\cmark$            & $\xmark$           & \multicolumn{1}{c|}{$0.35 / 0.95^{\phantom{\dagger}}$}                                          & 28.5                                  & 33.1                                  & 37.5                                  & 28.2                                  & 32.3                                  & 32.9                                  & 26.0                                  & \multicolumn{1}{c|}{18.7}                                  & 29.6                                  \\
            \multirow{-3}{*}{\phantom{ai}\textsc{ZeroSwot-Medium}} & \multicolumn{1}{r|}{(C)} & $\cmark$            & $\cmark$           & \multicolumn{1}{c|}{$0.35 / 0.95^\dagger$}                                                      & \cellcolor[HTML]{ECF4FF}30.5          & 34.9                                  & 39.4                                  & \cellcolor[HTML]{ECF4FF}30.6          & \cellcolor[HTML]{ECF4FF}35.0          & \cellcolor[HTML]{ECF4FF}37.1          & \cellcolor[HTML]{ECF4FF}27.8          & \multicolumn{1}{c|}{\cellcolor[HTML]{ECF4FF}20.3}          & \cellcolor[HTML]{ECF4FF}31.9          \\ \midrule
                                                                   & \multicolumn{1}{r|}{(A)} & $\xmark$            & $\xmark$           & \multicolumn{1}{c|}{$0.35 / 1.65^{\phantom{\dagger}}$}                                          & 26.5                                  & 31.1                                  & 33.5                                  & 25.4                                  & 29.9                                  & 30.6                                  & 24.3                                  & \multicolumn{1}{c|}{18.0}                                  & 27.4                                  \\
                                                                   & \multicolumn{1}{r|}{(B)} & $\cmark$            & $\xmark$           & \multicolumn{1}{c|}{$0.35 / 1.65^{\phantom{\dagger}}$}                                          & 30.1                                  & 34.8                                  & 38.9                                  & \cellcolor[HTML]{ECF4FF}29.8          & \cellcolor[HTML]{ECF4FF}34.4          & \cellcolor[HTML]{ECF4FF}35.3          & \cellcolor[HTML]{ECF4FF}27.6          & \multicolumn{1}{c|}{\cellcolor[HTML]{ECF4FF}20.4}          & \cellcolor[HTML]{ECF4FF}31.4          \\
            \multirow{-3}{*}{\phantom{ai}\textsc{ZeroSwot-Large}}  & \multicolumn{1}{r|}{(C)} & $\cmark$            & $\cmark$           & \multicolumn{1}{c|}{$0.35 / 1.65^\dagger$}                                                      & \cellcolor[HTML]{ECF4FF}\textbf{31.2} & \cellcolor[HTML]{ECF4FF}\textbf{35.8} & \cellcolor[HTML]{ECF4FF}40.5 & \cellcolor[HTML]{ECF4FF}\textbf{31.4} & \cellcolor[HTML]{ECF4FF}\textbf{36.3} & \cellcolor[HTML]{ECF4FF}\textbf{38.3} & \cellcolor[HTML]{ECF4FF}\textbf{28.0} & \multicolumn{1}{c|}{\cellcolor[HTML]{ECF4FF}\textbf{21.5}} & \cellcolor[HTML]{ECF4FF}\textbf{32.9} \\ \bottomrule \bottomrule
            \end{tabular}
            }
            \caption{BLEU($\uparrow$) scores on \textsc{MuST-C} \tstcommon. Upper part: In \textbf{bold} best among previous zero-shot ST and our proposed (small) model. Lower part: \underline{Underscored} is the best supervised score, \highlight{highlighted} are zero-shot scores that are better than the best supervised ones, and in \textbf{bold} is best overall. $^\dagger$ NLLB parameters are finetuned separately. $^*$ Parameters estimated from methodology description. Extended results in Table \ref{tab:extended_results_mustc}.}
            \label{tab:main_results_mustc}
        \end{table*}

        \textbf{\covost{}.} In Table \ref{tab:main_results_covost} we present the results of \zeroswot in \covost, where we used \cv for training, and a multilingually finetuned NLLB\footnote{Similar to scenario C of Table \ref{tab:data_usage}. Details in Appendix \ref{appendix:mt}.} on the MT data of \textsc{CoVoST} En-X. For comparison, we provide results from other supervised SOTA methods, which are XLS-R \cite{xls-r} and \seamless \cite{seamlessm4t,seamlessm4t-v2}, and group them according to their size based on inference parameters. For the medium-sized models, we observe that although our method is evaluated on zero-shot ST and has less training parameters, it surpasses the previous SOTA \seamless in 12/15 directions and on average by a large margin of 3.5 BLEU points. For the large models, \zeroswot ranks better than the original \lrg configuration of \seamless, with an average of 31.1 BLEU, outperforming it by 0.6 points. Compared to the updated v2, our method is 0.5 BLEU points behind, but still obtains better results in 7/15 directions.

        \begin{table*}[ht]
            \centering
            \resizebox{\textwidth}{!}{
            \begin{tabular}{@{}l|cc|ccccccccccccccc|c@{}}
            \toprule
            \multicolumn{1}{c|}{\textbf{Models}} & \textbf{ZS} & \textbf{Size (B)}  & \textbf{Ar}   & \textbf{Ca}   & \textbf{Cy}   & \textbf{De}   & \textbf{Et}   & \textbf{Fa}   & \textbf{Id}   & \textbf{Ja}   & \textbf{Lv}   & \multicolumn{1}{l}{\textbf{Mn}} & \textbf{Sl}   & \textbf{Sv}   & \textbf{Ta}   & \textbf{Tr}   & \textbf{Zh}   & \textbf{Average} \\ \midrule
            \multicolumn{19}{c}{\textit{Medium-sized models}}                                                                                                                                                                                                                                                                 \\
            XLS-R-1B              &       $\xmark$       & $1.0$               &  19.2          & 32.1          & \textbf{31.8} & 26.2          & 22.4          & 21.3          & 30.3          & 39.9          & 22.0          & 14.9                            & 25.4          & 32.3          & 18.1          & 17.1          & 36.7          & 26.0             \\
            \textsc{SeamlessM4T-M}     &    $\xmark$          & $1.2$               & 20.8          & 37.3          & 29.9          & \textbf{31.4} & 23.3          & 17.2          & 34.8          & 37.5          & 19.5          & 12.9                            & 29.0          & 37.3          & 18.9          & \textbf{19.8} & 30.0          & 26.6             \\ 
            \textsc{ZeroSwot-M (C)}    &   $\cmark$       & $0.35 / 0.95$        & \textbf{24.4} & \textbf{38.7} & 28.8          & 31.2          & \textbf{26.2} & \textbf{26.0} & \textbf{36.0} & \textbf{46.0} & \textbf{24.8} & \textbf{19.0}                   & \textbf{31.6} & \textbf{37.8} & \textbf{24.4} & 18.6          & \textbf{39.0} & \textbf{30.2}    \\ \midrule
            \multicolumn{19}{c}{\textit{Large-sized models}}                                                                                                                                                                                                                                                                  \\
            XLS-R-2B            &          $\xmark$          & $2.0$               &  20.7          & 34.2          & 33.8          & 28.3          & 24.1          & 22.9          & 32.5          & 41.5          & 23.5          & 16.2                            & 27.6          & 34.5          & 19.8          & 18.6          & 38.5          & 27.8             \\
            \textsc{SeamlessM4T-L}     &      $\xmark$       & $2.3$              &  24.5          & 41.6          & 33.6          & 35.9          & 28.5          & 19.3          & 39.0          & 39.4          & 23.8          & 15.7                            & 35.0          & 42.5          & 22.7          & 23.9          & 33.1          & 30.6             \\
            $\hookrightarrow$ v2    &       $\xmark$         & $2.3$               &  25.4          & \textbf{43.6} & \textbf{35.5} & \textbf{37.0} & \textbf{29.3} & 19.2          & \textbf{40.2} & 39.7          & 24.8          & 16.4                            & \textbf{36.2} & \textbf{43.7} & 23.4          & \textbf{24.7} & 35.9          & \textbf{31.7}    \\ 
            \textsc{ZeroSwot-L (C)}   &   $\cmark$         & $0.35 / 1.65$        &  \textbf{25.7} & 40.0          & 29.0          & 32.8          & 27.2          & \textbf{26.6} & 37.1          & \textbf{47.1} & \textbf{25.7} & \textbf{18.9}                   & 33.2          & 39.3          & \textbf{25.3} & 19.8          & \textbf{40.5} & 31.2             \\ \bottomrule
            \end{tabular}
            }
            \caption{BLEU($\uparrow$) scores on \covost En-X \test. ZS stands for zero-shot ST. For \zeroswot models, size is displayed in training/inference parameters. In \textbf{bold} are the best for each size category. Extended results in Table \ref{tab:extended_results_covost}.}
            \label{tab:main_results_covost}
        \end{table*}

    \textbf{How does \zeroswot compare with cascades?} To answer this, we train multilingual cascade ST systems based on \wv and NLLB in \textsc{MuST-C}. The cascades are trained in the same data scenarios and data sizes as \zeroswot (Tables \ref{tab:data_usage}, \ref{tab:model_size}), and have two versions based on the ASR model, which is either a CTC encoder or an attention-based encoder-decoder (AED), in which case we pair \wv with a Transformer decoder (Appendix \ref{appendix:cascade}). We present BLEU scores by data usage and model size, and also measure the average inference time per example. The results of Table \ref{tab:cascade_mustc_results} highlight that although \zeroswot closely resembles a CTC-based cascade, in that it softly connects a CTC encoder and an MT model, its performance is consistently better, with gains from 2.7 to 7.2 BLEU points\footnote{The improvements are more evident than the zero-shot ST method of \citet{towards-zero}, which surpassed CTC-based cascades by 0.8 BLEU, further highlighting the effectiveness of our method complimented by the proposed compression.}. This can be expected since CTC ASR models are error-prone, leading to compounding errors through the MT model \cite{cascade_vs_direct}, where our method solves this by directly mapping the encoder representation to the MT embedding space via OT and our proposed compression. Compared to strong cascades that utilize AED-based ASR models, we observe that \zeroswot is noticeably better in the absence of in-domain data (A, +0.8/0.9), while the cascade becomes slightly better when in-domain data become fully available (C, -0.3/0.3). More importantly though, we find that \zeroswot excibits better efficiency, being $18\%$ and $5\%$ faster in the \med and \lrg models accordingly.

        \begin{table}[ht]
            \centering
            \resizebox{\columnwidth}{!}{
            \begin{tabular}{@{}lcccccccc@{}}
            \toprule
            \multicolumn{1}{l|}{}                              & \multicolumn{6}{c|}{\textbf{BLEU}}                                                                                                                                                                                                                                                                                                             & \multicolumn{2}{c}{\textbf{Time}}                                                                              \\ \midrule
            \multicolumn{1}{l|}{\textbf{Scenario}}             & \multicolumn{2}{c}{\textbf{A}}                                                                                & \multicolumn{2}{c}{\textbf{B}}                                                                                & \multicolumn{2}{c|}{\textbf{C}}                                                                                & \multicolumn{2}{c}{\textbf{/}}                                                                                 \\ \midrule
            \multicolumn{1}{l|}{\textbf{Model Size}}           & \textbf{M}                                            & \textbf{L}                                            & \textbf{M}                                            & \textbf{L}                                            & \textbf{M}                                            & \multicolumn{1}{c|}{\textbf{L}}                        & \textbf{M}                                             & \textbf{L}                                            \\ \midrule
            \multicolumn{9}{c}{\textit{Cascades with \wv based ASR model and NLLB}}                                                                                                                                                                                                                                                                                                                                                                                                                                              \\
            \multicolumn{1}{l|}{CTC-based}                     & 21.8                                                  & 24.7                                                  & 25.4                                                  &         27.5                                              & 24.7                                                  & \multicolumn{1}{c|}{26.5}                              & \textbf{0.41}                                          & \textbf{0.66}                                         \\
            \multicolumn{1}{l|}{AED-based}                     & 25.2                                                  & 26.5                                                  & \textbf{30.0}                                         & 31.1                                                  & \textbf{32.2}                                         & \multicolumn{1}{c|}{\textbf{33.2}}                     & 0.56                                                   & 0.82                                                  \\ \midrule
            \multicolumn{1}{l|}{\zeroswot}                     & \textbf{26.0}                                         & \textbf{27.4}                                         & 29.6                                                  & \textbf{31.4}                                         & 31.9                                                  & \multicolumn{1}{c|}{32.9}                              & 0.46                                                   & 0.78                                                  \\
            \multicolumn{1}{l|}{$\hookrightarrow$ $\Delta$ CTC} & \multicolumn{1}{r}{\color[HTML]{3531FF} \small+$4.2$} & \multicolumn{1}{r}{\color[HTML]{3531FF} \small+$2.7$} & \multicolumn{1}{r}{\color[HTML]{3531FF} \small+$4.2$} & \multicolumn{1}{r}{\color[HTML]{3531FF} \small+$3.9$} & \multicolumn{1}{r}{\color[HTML]{3531FF} \small+$7.2$} & \multicolumn{1}{r|}{\color[HTML]{3531FF} \small+$6.4$} & \multicolumn{1}{r}{\color[HTML]{FD6864} \small+$12\%$} & \multicolumn{1}{r}{\color[HTML]{FD6864} \small+$18\%$} \\ 
            \multicolumn{1}{l|}{$\hookrightarrow$ $\Delta$ AED} & \multicolumn{1}{r}{\color[HTML]{3531FF} \small+$0.8$} & \multicolumn{1}{r}{\color[HTML]{3531FF} \small+$0.9$} & \multicolumn{1}{r}{\color[HTML]{FD6864} \small-$0.4$} & \multicolumn{1}{r}{\color[HTML]{3531FF} \small+$0.3$} & \multicolumn{1}{r}{\color[HTML]{FD6864} \small-$0.3$} & \multicolumn{1}{r|}{\color[HTML]{FD6864} \small-$0.3$} & \multicolumn{1}{r}{\color[HTML]{3531FF} \small-$18\%$} & \multicolumn{1}{r}{\color[HTML]{3531FF} \small-$5\%$} \\ \bottomrule
            \end{tabular}
            }
            \caption{Average BLEU($\uparrow$) scores and average inference Time($\downarrow$) in seconds on \textsc{MuST-C} \tstcommon. M/L the size of NLLB for each model (600M / 1.3B). Extended results in Table \ref{tab:extended_results_mustc}.}
            \label{tab:cascade_mustc_results}
            \vspace{-0.3cm}
        \end{table}
        \begin{table}[ht]
            \centering
            \resizebox{\columnwidth}{!}{
            \begin{tabular}{@{}cccccc@{}}
            \toprule
            \textbf{Compression} & \textbf{Tokenization} & \textbf{BLEU} & $\boldsymbol{|\Delta \text{len}|}$ & $r(\text{len})$ \\ \midrule
            None                 & Word                                &  28.9             & 267.6       &  10.9       \\
            Length Adaptor       & Word                                &  28.9             & 49.2   & 2.82           \\
            Character-level      & Word                                &  29.0             & 71.2    & 3.61           \\
            Compr. Adapter          & Word                                &  27.2             & 6.2   & 0.77            \\
            Compr. Adapter         & Subword                             &  28.4             & 3.3    & 0.88           \\ \midrule
            Compr. Adapter          & Subword + Unk                       & \textbf{29.6}     & \textbf{1.4}    & \textbf{0.98}        \\ \bottomrule
            \end{tabular}
            }
            \caption{Average BLEU($\uparrow$) scores, average absolute length difference $\boldsymbol{|\Delta \text{len}|}$($\downarrow$), and average length ratio $r(\text{len})$ (closer to 1 is better), between compressed speech and text representations. Results with \zeroswot-\med(B) on \textsc{MuST-C} \tstcommon. \emph{Compression Adapter} and \emph{Subword + Unk} is our proposal. Extended results in Table \ref{tab:extended_results_mustc}.}
            \label{tab:compression_results}
            \vspace{-0.3cm}
        \end{table}
            
        \textbf{Does our compression reduce the length gap?} We replace our proposed compression adapter (\S \ref{subsec:meth.compress}), with either a length adaptor \cite{lna} that down-samples by a factor of 4, or CTC compression \cite{ctc_compression}, or no compression at all.\footnote{Details on these experiments can be found in App. \ref{appendix:zeroswot_compression}.} We also consider different types of tokenization than the proposed "Subword+Unk" (Table 
        \ref{tab:ctc_labels}), which affect the placement of the separator tokens in the CTC target labels, and thus which representations are combined by the compression adapter. Apart from translation quality in \textsc{MuST-C}, we measure the length gap between the speech and text representations at the output of the shared encoder, in terms of absolute differences $\boldsymbol{|\Delta \text{len}|} = |\text{len}(speech) - \text{len}(text)|$, and relative ratios $r(\text{len}) = \text{len}(speech) / \text{len}(text)$. Our findings in Table \ref{tab:compression_results} show that our proposed method (final row) surpasses previous methods (rows 1-3), by 0.6-0.7 points, and closes the absolute length gap to an average of 1.4 tokens. We also find that without learning the text branch tokenization (rows 4-5), our compression falls behind previous methods. The length ratio of each method reveals that \emph{over-compression}, i.e. $r(\text{len}) < 1$, is what worsens performance, leading to information loss, as even the model without compression ($r(\text{len}) \! \approx \! 11$) achieves scores higher than our method without proper tokenization with respect to the text branch.


        \textbf{Does \zeroswot close the modality gap?} Apart from the competitive zero-shot performance we provide further evidence by designing a cross-modal retrieval experiment. In Table \ref{tab:retrieval_accuracy} we present the speech-text retrieval accuracy in the 2551 examples of \textsc{MuST-C} En-De \tstcommon, and compare against ConST \cite{const}, which used contrastive learning. We perform two retrieval experiments. For each speech representation we retrieve the text representation with the (a) highest cosine similarity after temporal mean-pooling or (b) the lowest Wasserstein distance. Our results show that \zeroswot with Wasserstein-based retrieval achieves an improvement of 3.5 points compared to ConST, approaching a perfect accuracy score\footnote{Manual inspection of the 32 wrong predictions, showed that most of them are either due to many positives (several "Thank you.") or noisy examples (misaligned speech and text).}, which highlights the degree of alignment between the text and the learned speech representations.

        \begin{table}[ht]
            \centering
            \resizebox{0.65\columnwidth}{!}{
            \begin{tabular}{@{}lc@{}}
            \toprule
            \textbf{Model}                    & \textbf{Accuracy} \\ \midrule
            ConST \cite{const}               & 95.0  \\
            \textsc{ZeroSwot} (Cosine)              & 98.1  \\
            \textsc{ZeroSwot} (Wass)   & \textbf{98.5} \\ \bottomrule
            \end{tabular}
            }
            \caption{Speech-to-Text retrieval accuracy($\uparrow$) on \textsc{MuST-C} En-De \tstcommon. Results with \zeroswotm (B).}
            \label{tab:retrieval_accuracy}
            \vspace{-0.2cm}
        \end{table}

        \textbf{Is \zeroswot massively multilingual?} We evaluate on FLEURS \cite{fleurs}, from English to 88 target languages. In Table \ref{tab:fleurs} we present the BLEU scores of \zeroswot trained on \cv using the original NLLB versions. Since our model can benefit from ASR data which are in general accessible for English, we additionally train with \mustcasr and \ls. We also compare with a strong cascade by training an AED-ASR model based on \wv (similar to Table \ref{tab:cascade_mustc_results}) on \cv and coupling it with the original NLLB. Although our models are zero-shot ST systems, we find that they obtain results that are competitive to \seamless, which used 400k hours of audio data \cite{seamlessm4t}. Furthermore \zeroswot is comparable to cascades, and additional ASR data can bring some further improvements.

        \begin{table}[ht]
            \centering
            \resizebox{\columnwidth}{!}{
            \begin{tabular}{@{}lccc@{}}
            \toprule
            \textbf{Model} & \textbf{Type}  & \textbf{Medium} & \textbf{Large} \\ \midrule
            NLLB   & MT               & 22.5       & 24.8       \\ \midrule
            AED-ASR + NLLB & Cascade-ST                 & 17.8       & 20.2       \\
            \textsc{ZeroSwot} & ZS-ST          & 18.1       & 20.1       \\
            $\hookrightarrow$ more data & ZS-ST & 18.4       & 20.3       \\
            \textsc{SeamlessM4T}  & E2E-ST     & \textbf{19.2} & \textbf{21.5} \\ \bottomrule
            \end{tabular}
            }
            \caption{Average BLEU($\uparrow$) scores on FLEURS \test. In \textbf{bold} is best ST scores. Extended results in Table \ref{tab:extended_results_fleurs}.}
            \label{tab:fleurs}
        \end{table}

    \textbf{Can \zeroswot benefit from ST finetuning?} We initialize ST models with the zero-shot models trained on \textsc{MuST-C} (Table \ref{tab:main_results_mustc}) with medium/large sizes and B/C data usage scenarios, and perform supervised ST finetuning on \textsc{MuST-C} En-De (Appendix \ref{appendix:zeroswot_st_finetuning}). The results of Table \ref{tab:st_finetuning} hint that \zeroswot when trained on both ASR and MT data (scenario C) is likely in a local minima, which is not easily adaptable to supervised ST, having only marginal improvements (+0.2 BLEU). On the contrary, models trained only with ASR \textsc{MuST-C} data (scenario B) are more flexible for adaptation, with gains of 1.9-2.3 BLEU points. Notably, by finetuning \zeroswot-\lrg (B) we obtain an even higher new SOTA for \textsc{MuST-C} En-De at 32 BLEU points, improving the previously best published result \cite{lst} by 1.6 points (Table \ref{tab:main_results_mustc}).
    
    \begin{table}[ht]
        \centering
        \resizebox{\columnwidth}{!}{
        \begin{tabular}{@{}ccccc@{}}
        \toprule
        \multicolumn{2}{c}{\textbf{Model Type}} & \multicolumn{3}{c}{\textbf{BLEU}} \\ 
        \cmidrule(lr){1-2} \cmidrule(l){3-5}
        \textbf{Size} & \textbf{Scenario} & \textbf{Zero-shot} & \textbf{Finetuned} & $\mathbf{\Delta
        }$ \\ 
        \midrule
        \med & B & 28.5 & 30.8 & {\color[HTML]{00009B} \small{+$2.3$}} \\
        \med & C & 30.5 & 30.7 & {\color[HTML]{3B9DB9} \small{+$0.2$}} \\
        \lrg & B & 30.1 & \textbf{32.0} & {\color[HTML]{00009B} \small{+$1.9$}} \\
        \lrg & C & 31.2 & 31.4 & {\color[HTML]{3B9DB9} \small{+$0.2$}} \\ 
        \bottomrule
        \end{tabular}
        }
        \caption{BLEU($\uparrow$) scores on \textsc{MuST-C} En-De \tstcommon with zero-shot training and with supervised ST finetuning. In \textbf{bold} is best overall.}
        \label{tab:st_finetuning}
    \end{table}

    \textbf{Ablations.} In the ablations of Table \ref{tab:ablations} we observe that the speech embedder (\S \ref{sec:meth.speech_embedder}) is critical to the effectiveness of \textsc{ZeroSwot}, possibly due to reasons regarding over-compression (absence of \texttt{<lang>} and \texttt{</s>}). Auxiliary Wasserstein losses (\S \ref{subsec:meth.final_loss}) have less impact, but come with a negligible computational cost due to batching.

    \begin{table}[ht]
        \centering
        \resizebox{0.65\columnwidth}{!}{
        \begin{tabular}{@{}lc@{}}
        \toprule
        \textbf{Model}                       & \textbf{BLEU} \\ \midrule
        \textsc{ZeroSwot-Medium} (B)                                & \textbf{29.6}         \\
        $\hookrightarrow$ w/o Speech Embedder               &  28.6          \\
        $\hookrightarrow$ w/o Auxiliary Wass. & 29.5 \\ \bottomrule
        \end{tabular}
        }
        \caption{Avg BLEU($\uparrow$) on \textsc{MuST-C} \tstcommon.}
        \label{tab:ablations}
        \vspace{-0.3cm}
    \end{table}

\setlength{\parindent}{12pt}

\section{Conclusions} \label{sec:conclusions}

    We presented \zeroswot, a novel method for zero-shot end-to-end Speech Translation, that works by adapting a speech encoder to the representation space of a massively multilingual MT model. Our method sets new SOTA in \text{MuST-C}, while it also surpasses larger, supervised models in \textsc{CoVoST}. We additionally showed that \zeroswot outperforms cascades, both in performance and efficiency. Finally, we provided evidence of closing the modality gap, with respect to both length and representation. Future research will expand on low-resource scenarios and speech-to-speech translation.
    
\section*{Limitations}

    Our method obtains state-of-the-art results in Speech Translation, despite being a zero-shot model. Still there are a couple of limitations inherent to the model, and on its evaluation. Although \zeroswot is an end-to-end model, in the sense that there is not intermediate transcription step and solves error-propagation, it suffers from some of the caveats of cascade systems. Since the speech encoder "mimics" the representation space of an MT model, no acoustic information is retained. This means that the resulting translation model can't use acoustic information, like prosody, and can thus be sometimes limited in each ability to carry out correct translations. Furthermore, given the requirement of ASR data to train the speech encoder, this methodology can't be used for spoken-only languages, at least in its current form. 
    
    Finally, in our evaluations we did not test for a different language, other than English, in the source. This was partly due to space constrains in the paper. Although we hypothesize that our proposed method can handle equally well other source languages, it would be interesting to be applied for low-resource scenarios, which we leave as future research.


\section*{Acknowledgements}

Work at UPC was supported by the Spanish State Research Agency (AEI) project PID2019-107579RB-I00 / AEI / 10.13039/501100011033.

\bibliography{anthology,custom}
\bibliographystyle{acl_natbib}

\appendix

\section{Data Processing and Filtering} \label{appendix:data}

    \begin{table}[ht]
        \centering
        \resizebox{0.9\columnwidth}{!}{
        \begin{tabular}{@{}lcccc@{}}
        \toprule
        \multirow{2}{*}{\textbf{Dataset}} & \multicolumn{2}{c}{\textbf{Size}}  & \multicolumn{2}{c}{\textbf{Average Length}} \\ \cmidrule(l){2-5} 
                                          & \textbf{examples} & \textbf{hours} & \textbf{speech}       & \textbf{text}       \\ \midrule
        \ls                       & 281K              & 961            & 12.3                  & 46.6                \\
        \cv                       & 949K              & 1,503          & 5.7                   & 17.2                \\
        \mustcasr                        & 352K              & 681            & 7.0                   & 29.7                \\ \bottomrule
        \end{tabular}
        }
        \caption{Dataset statistics for Zero-shot ST training. Speech length is measured in seconds, and text length in tokens.}
        \label{tab:data}
    \end{table}
    
    \subsection{\mustc}
    
        For the \train set of each language pair, we perform text cleaning on both the transcription and translation, including space and punctuation normalization as well as removal of non-utf characters. We also remove speaker names and events such as "(Laughing)". Following, we remove examples that contain less than 4 characters, or have a transcription-to-translation ratio smaller than 0.5 or larger than 2. For the Zero-shot ST training we remove the translations, and process the transcriptions by replacing numbers and symbols with their spelled out forms. Then we concatenate all eight training sets and remove duplicates according to the triplet of (\emph{speaker}, \emph{text}, \emph{duration}). Due to alignment errors during the creation of the dataset, the same text can be paired with different parts of audio for different language pairs. Thus, for duplicates of (\emph{speaker}, \emph{text}) we keep the example that is closer to the speaking ratio of speaker\footnote{Calculated as the average words-per-second of all the examples for this speaker in the concatenated dataset.}. Finally, we remove examples with extreme words-per-second ratio (larger than 10). A summary of the dataset sizes is available at Table \ref{tab:mustc_data}. We also similarly construct a validation set, with a total of 2,081 examples, to be used during training, by concatenating the individual {\fontfamily{qcr}\selectfont dev} sets (without filtering), and simple removal of hard duplicates according to the triplet of (\emph{speaker}, \emph{text}, \emph{duration}).

        \begin{table}[ht]
            \centering
            \resizebox{\columnwidth}{!}{
            \begin{tabular}{@{}lcc@{}}
            \toprule
            \textbf{Dataset}                           & \textbf{Examples (K)} & \textbf{Hours} \\ \midrule
            En-De                                      & 220                   & 384            \\
            En-Es                                      & 254                   & 473            \\
            En-Fr                                      & 262                   & 466            \\
            En-It                                      & 242                   & 438            \\
            En-Nl                                      & 237                   & 416            \\
            En-Pt                                      & 196                   & 359            \\
            En-Ro                                      & 226                   & 406            \\
            En-Ru                                      & 250                   & 458            \\ \midrule
            \mustcasr                                   & 1,888                 & 3,402          \\
            $\hookrightarrow$ Duplicate removal        & 367                   & 713            \\
            $\hookrightarrow$ Speaking ratio filtering & 353          & 681   \\ \bottomrule
            \end{tabular}
            }
            \caption{Number of examples (in thousands) and total duration of \mustc{ } \train set used to train \zeroswot.}
            \label{tab:mustc_data}
        \end{table}

    \subsection{\cv}

        We use version 8.0\footnote{\href{https://commonvoice.mozilla.org/en/datasets}{commonvoice.mozilla.org/en/datasets}}, and more specifically the {\fontfamily{qcr}\selectfont train} and {\fontfamily{qcr}\selectfont dev} splits. For the {\fontfamily{qcr}\selectfont train} split, we apply text normalization, as well as replace numbers and symbols with their spelled-out forms.

    \subsection{LibriSpeech}

        For training we use the {\fontfamily{qcr}\selectfont train-clean-100}, {\fontfamily{qcr}\selectfont train-clean-360}, and {\fontfamily{qcr}\selectfont train-other-500} splits, and for evaluation the {\fontfamily{qcr}\selectfont dev-clean} one. Since LibriSpeech has normalized transcriptions, we restore the casing and punctuation with a finetuned BERT-base model \cite{bert} using rpunct.\footnote{\href{https://github.com/Felflare/rpunct}{github.com/Felflare/rpunct}}

    \subsection{CTC Labels}

        For all the datasets, to obtain the target labels for CTC, first we tokenize the transcription with the NLLB sentencepiece model\footnote{\href{https://tinyurl.com/flores200sacrebleuspm}{fairseq/nllb/flores200\_sacrebleu\_tokenizer\_spm.model}}, then remove casing, replace all spaces with the \texttt{<sep>} token, and substitute all characters not present in the \wv letter-based vocabulary\footnote{\href{https://dl.fbaipublicfiles.com/fairseq/wav2vec/dict.ltr.txt}{fairseq/wav2vec/dict.ltr.txt}} with the \texttt{<unk>} token. 
    
\section{Evaluation} \label{appendix:evaluation}

    The evaluation of all models, either MT, \zeroswot, ASR, Cascade ST, or Supervised ST remains the same. We apply checkpoint averaging to the 10 best checkpoints according to the corresponding validation metric, and generate with a beam search of 5. We measure translation quality with BLEU \cite{bleu} using sacreBLEU \cite{sacrebleu} with signature \texttt{BLEU|nrefs:1|case:mixed|eff:no|tok:13a|}
    \texttt{smooth:exp|version:2.3.1}. For comparing with \seamless, we follow their evaluation, and using a \texttt{char} tokenization for Mandarin Chinese, Japanese, Thai, Lao and Burmese. We do not carry statistical significance testing since in almost all results we are comparing average BLEU scores across many language directions. Although BLEU as a metric has its limitations \cite{bleu_problems}, we opted to use it for comparison with previous methods from the literature.

\section{\zeroswot Models} \label{appendix:zeroswot}

    \subsection{Model Architecture} \label{appendix:zeroswot_architecture}
    
        The acoustic encoder is composed of a feature extractor with 7 strided convolutional layers that sub-sample the signal by $r\!=\!320$, followed by $N\!=\!24$ transformer layers with $d\!=\!1024$ \cite{wav2vec2.0}. Each layer has feed-forward dimension of 4096, 16 attention heads, GELU activations \cite{gelu}, and use the pre-layernorm configuration \cite{prelayernorm}. The acoustic encoder and the linear layer of the CTC with size $|\mathcal{V}|\!=\!30$ are initialized from a large version of \wv\footnote{\href{https://dl.fbaipublicfiles.com/fairseq/wav2vec/wav2vec_vox_960h_pl.pt}{fairseq/wav2vec/wav2vec\_vox\_960h\_pl.pt}}, with around 315M parameters, which was pretrained with self-supervised learning and self training \cite{wav2vec-selftraining}, and finetuned on ASR with Librispeech \cite{librispeech} and Libri-light \cite{librilight}.
        
        The subword encoder in the compression adapter is a Transformer encoder with 3 layers and dimensionality $d\!=\!1024$, which have the same format as the ones in the acoustic encoder. We also add sinusoidal positional encodings to the chunked input.
        
        The text embedding uses a vocabulary with size $|\mathcal{B}|\!=\!256k$, and for the text encoder we are either using the \med version with $L\!=\!12$ layers or the \lrg version with $L\!=\!24$ layers. Each layer in the \med version has a dimensionality of 1024, feed-forward dimension of 4096, ReLU activation, 16 attention heads, and pre-layernorm configuration. The only difference of the \lrg version apart from the number of layers is the feed-forward dimension which is 8192. For initialization we are using two dense versions of NLLB \cite{nllb}, which have been distilled from the 54B parameter sparse model: the 600M version\footnote{\href{https://tinyurl.com/nllb200densedst600mcheckpoint}{fairseq/nllb/nllb-200-distilled-600M.pt}} and the 1.3B version\footnote{\href{https://tinyurl.com/nllb200densedst1bcheckpoint}{fairseq/nllb/nllb-200-distilled-1.3B.pt}}.

        The \textsc{Small} version of our method uses the \textsc{Base}\footnote{\href{https://dl.fbaipublicfiles.com/fairseq/wav2vec/wav2vec_small_960h.pt}{fairseq/wav2vec/wav2vec\_small\_960h\_pl.pt}} version of \wv (12 layers and dimensionality 768) and NLLB-\med. The subword encoder has 3 layers, with dimensionality 768. We use a linear layer to project its output to the $d=1024$, which is the dimensionality of the text encoder due to NLLB-\med. The total number of training parameters is 110M, while the inference parameters are approximately 0.7B.

    \subsection{Training details} \label{appendix:zeroswot_training}
    
        We finetune the parameters of the acoustic encoder, CTC module and compression adapter, while the parameters of the speech embedder, text embedding and text encoder are kept frozen. We are using AdamW (0.9, 0.98) \cite{adamw} with a base learning rate of 3e-4, an inverse square root scheduler with a linear warmup (2k for \mustcasr, 4k for \cv), and a batch size 32M tokens. The input to the speech branch is raw waveforms sampled at 16kHz and the input to the text branch is English text, tokenized with the NLLB sentencepiece model\footnote{\href{https://tinyurl.com/flores200sacrebleuspm}{fairseq/nllb/flores200\_sacrebleu\_tokenizer\_spm.model}}.

        For the Wasserstein loss we are using the geomloss package\footnote{\href{https://www.kernel-operations.io/geomloss}{kernel-operations.io/geomloss}} using the Sinkhorn distance \cite{sinkhorn} with the default parameters. When it for multiple layers, we batch them together for efficient computation. The Wasserstein losses for the intermediate layers are applied after the first layer-norm of the next layer, since we are using a pre-layernorm setting. Applying the losses on representations that have not been layernormed made the training unstable.
        
        We use an activation dropout of 0.1 in the acoustic encoder, a dropout of 0.1 in the CTC module, and a dropout of 0.1 in the compression adapter. We furthermore apply masking to the speech signal \cite{wav2vec2.0}, with a probability of 0.3 to mask 10 consecutive frames, and a probability of 0.2 to mask 64 consecutive frequency channels. 
        
        To speed-up the training, we extract offline the text encoder representations from NLLB, and can thus completely remove the frozen text branch during the training of the speech encoder.
        
        We train until convergence, and apply early stopping if the Wasserstein loss of the last layer in the development set using has not improved for 10 epochs. Models on \ls or \mustcasr usually train for around 50k steps, while on \cv they train for around 150k.
        
        To accelerate the training for models in \cv, we trained a \emph{seed} model using the original \med version of NLLB, and then adapted the resulting speech encoder to different NLLB versions (\lrg and/or finetuned). For adaptation, only required 30k steps (compared to 150k of the seed model), where we used learning of 3e-5, warm-up of 1k steps and cosine annealing. 

         All models are implemented and trained on \textsc{Fairseq} \cite{fairseq,fairseq-s2t} with \texttt{memory-efficient-fp16}. Training for 50k steps takes around 36 hours on a machine with 8 NVIDIA 3090s (for either \med or \lrg).

    \subsection{Models trained with different type of compression} \label{appendix:zeroswot_compression}

        Here we provide details for the models trained to obtain the results of Table \ref{tab:compression_results}. To ensure a fair comparison we keep the same number of parameters for all models, by adding transformer layers or MLPs. Architecture changes are done only in the part of the compression, and everything else remains the same (including Speech Embedder). For the model without compression we add two large MLPs with residual connections at the output of the acoustic encoder. The MLPs consist of up-projection to 8192, GELU activation, and down-projection back to 1024. For the model with the Length Adaptor \cite{lna}, we use 2 strided convolutional layers that sub-sample the output of the acoustic encoder by a factor of 4, followed by a transformer layer. For the model using CTC-based compression \cite{bridging_the_modality_gap,ctc_compression} we first mean-pool consecutive same-labeled representations as in \S \ref{subsec:meth.compress}, followed by 3 transformer layers. For the models using the proposed compression adapter but different tokenizations, we just change the CTC target labels, and everything else remains the same.

    \subsection{Supervised ST finetuning} \label{appendix:zeroswot_st_finetuning}

        Here we provide details for the ST-finetuned versions of \zeroswot from Table \ref{tab:st_finetuning}. We train for 20k steps with a learning rate of 4e-5, linear warmup for 1k steps, fixed for another 4k steps, and reduced with cosine annealing for the rest. For \med models we finetune the parameters of the compression adapter, text encoder and text decoder, and thus keep frozen the acoustic encoder, speech embedder, and text embedding. For the \lrg models, we only finetune the compression adapter and apply LNA to the text encoder-decoder \cite{lna}. We hypothesize that better results could be obtain with more sophisticated finetuning approaches \cite{lora}. For inference we average the 10 best checkpoints according to BLEU on \textsc{MuST-C} En-De \texttt{dev}. 
        
\section{Machine Translation Models} \label{appendix:mt}

    \setlength{\parindent}{0pt}

    \textbf{Model Architecture.} For the MT models used in this research we used NLLB-600M (\med) and NLLB-1.3B (\lrg), which are dense transformers distilled from a 54B parameter Mixture-of-Experts (MoE) model \cite{nllb}. NLLB supports many-to-many translation between 202 languages. The \med model has 12 layers in both encoder and decoder, model dimensionality of 1024, feed-forward dimension of 4096, 16 attention heads, ReLU activations, and using pre-layernorm \cite{prelayernorm}. The only difference of the \lrg model is that has 24 layers in both encoder and decoder, and a feed-forward dimension of 8192. Both use a multilingual shared vocabulary of 256k tokens learned with sentencepiece \cite{sentencepiece}.

    \textbf{Finetuning hyperparameters.} In order to adapt them to the domain and languages of \textsc{MuST-C} and \covost, we did multilingual finetuning of NLLB using their transcription-translation pairs. We used batch size of 32k tokens for the \med, and 64k for \lrg. We used the AdamW optimizer \cite{adamw} with 2k warm-up updates and an inverse square root scheduler. The learning rate and dropout rate were selected from a grid search to \{7.5e-5, 1e-4, \textbf{2e-4}\} and \{0.1, \textbf{0.2}, 0.3\} accordingly\footnote{For each NLLB version we run 9 experiment in \textsc{MuST-C} and used the optimal combinations also for \covost.}. We also use a dropout of 0.1 for the attention, 0.0 for the activations and a label smoothing of 0.1 for the cross entropy loss, as was done for training the original versions \cite{nllb}. We evaluate every 500 steps on the {\fontfamily{qcr}\selectfont dev} sets, and early stop if the loss has not improved for 20 consecutive evaluations. At the end of the training we average the best 10 checkpoints according to the {\fontfamily{qcr}\selectfont dev} sets loss. Models were trained with \texttt{fp16}, while to train the \lrg version on an NVIDIA 3090, we used \texttt{memory-efficient-fp16} \cite{fairseq}. Models in \textsc{MuST-C} required 40k steps to converge, while the ones in \covost required around 100k.

    \textbf{Results.} BLEU scores for both the original and finetuned versions are available at Table \ref{tab:extended_results_mustc} for \textsc{MuST-C} and Table \ref{tab:extended_results_covost} for \covost.

    \setlength{\parindent}{12pt}
    
\section{Cascade Speech Translation Models} \label{appendix:cascade}

    Here we provide the training details and hyperparameters used for the Cascade ST systems (Table \ref{tab:cascade_mustc_results}). For the MT models we used either NLLB-\med or NLLB-\lrg, which were optionally finetuned, as described in Appendix \ref{appendix:mt}. For the ASR models we either used CTC or attention-based encoder-decoder (AED) models, as described below.

    \setlength{\parindent}{0pt}

    \textbf{CTC ASR Models.} The architecture is the same with the acoustic encoder used in the \zeroswot models, followed by a CTC module (Appendix \ref{appendix:zeroswot}). We initialize the model with the same version used to initialize the \zeroswot models, and finetune it with CTC on either \textsc{MuST-C}(ASR)or \cv. For each dataset, we use exactly the same filtering and preprocessing (Appendix \ref{appendix:data}), and as targets for the CTC we use the standard labeling (Table \ref{tab:ctc_labels}, 1st row). We use AdamW \cite{adamw} with a learning rate of 1e-4, warmup of 1k steps and an inverse square root scheduler, and the batch size is set to 32M tokens. We use a dropout of 0.1 for the activations and at the CTC module, and a time masking with probability 0.5 for length of 10, and channel masking with probability of 0.25 for size of 64 \cite{wav2vec2.0}. We evaluate every 250 steps and stop the training when the validation loss did not improve for 10 consecutive evaluations. We average the 10 best checkpoints according to the validation loss. The size of the model is around 315 million parameters. Results for these models are available at Table \ref{tab:asr_results_mustc}.

    \textbf{AED ASR Models.} The architecture of the encoder is the same as with the CTC ASR model. For the decoder we use 6 transformer layers with dimensionality of 768, feed-forward dimension of 3072, 8 attention heads, GeLU activations \cite{gelu}, and pre-layernorm \cite{prelayernorm}. We initialize the encoder with a pretrained \wv\footnote{\href{https://dl.fbaipublicfiles.com/fairseq/wav2vec/wav2vec_vox_new.pt}{fairseq/wav2vec/wav2vec\_vox\_new.pt}}, train the whole model with cross entropy using \ls and then finetune it either on \mustcasr or \cv. The target vocabulary has a size of 16k and is learned jointly on the three ASR datasets with sentencepiece. For each dataset, we use exactly the same filtering and preprocessing (Appendix \ref{appendix:data}), and the target text is the same we used for training the \zeroswot models. The hyperparameters are the same for both training on \ls and finetuning on \mustcasr or \cv. We use AdamW \cite{adamw} with learning rate of 2e-4, an inverse square root scheduler with a warmup of 1k steps, and a batch size of 32M tokens. The masking and dropout for the encoder are as in the CTC model, while for the decoder we use a dropout of 0.1, and a label smoothing of 0.1. We evaluate every 500 steps and stop training when the validation loss did not improve for 10 consecutive updates. We average the 10 best checkpoints according to the validation loss. The size of the model is around 390 million parameters. Results for these models are available at Table \ref{tab:asr_results_mustc}.

    \textbf{Cascade ST inference.} For generation, we use a beam search of size 5 for either the CTC or AED ASR model, and then pass the ASR generated text to the MT model, which generates the translation with a beam search of size 5.

    \setlength{\parindent}{12pt}

\section{Detailed Results in \mustc} \label{appendix:results_mustc_full}

    In Table \ref{tab:extended_results_mustc} we provide a summary of all obtained results in \textsc{MuST-C}, with our \zeroswot models (including ablations), our cascade models, and our finetuned NLLB models. We also provide more results from the literature for both zero-shot and supervised ST models. Note on performance drop of CTC-based Cascade when moving from the original NLLB model (scenario B) to the finetuned one (scenario C). We hypothesize that the original NLLB can better handle the noisy English text produces by the CTC ASR model, due to the abundance of data it was trained on. When finetuning it using the bitext of \textsc{MuST-C}, the model losses some of this capacity.

    In Table \ref{tab:asr_results_mustc} we provide the results in \textsc{MuST-C} of the ASR models we trained for the CTC-based and AED-based cascades (Tables \ref{tab:cascade_mustc_results}, \ref{tab:extended_results_mustc}). Models trained on \cv  correspond to scenario (A), while models trained on \textsc{MuST-C} correspond to scenarios B and C. 

    \begin{table*}[ht]
        \centering
        \resizebox{\textwidth}{!}
        {
        \begin{tabular}{@{}lrcccccccccccc@{}}
        \toprule \toprule
        \multicolumn{2}{c}{\multirow{2}{*}{\textbf{Models}}}                                                                                                            & \multicolumn{2}{c}{\textbf{MuST-C Data}} & \multirow{2}{*}{\textbf{\begin{tabular}[c]{@{}c@{}}Training/Inferece\\ Params (B)\end{tabular}}} & \multicolumn{9}{c}{\textbf{BLEU}}                                                                                                              \\ \cmidrule(lr){3-4} \cmidrule(l){6-14} 
        \multicolumn{2}{c}{}                                                                                                                                            & \textbf{ASR}        & \textbf{MT}        &                                                                                                  & \textbf{De} & \textbf{Es} & \textbf{Fr} & \textbf{It} & \textbf{Nl} & \textbf{Pt} & \textbf{Ro} & \textbf{Ru}               & \textbf{Average} \\ \midrule 
        \multicolumn{14}{c}{\textit{Machine Translation}}                                                                                                                                                                                                                                                                                                                                                                                                       \\
        \multicolumn{2}{l|}{NLLB-\med (original, used in A, B)}                                                                                                                       & -                   & $\xmark$           & \multicolumn{1}{c|}{0.6}                                                                         & 32.7        & 36.9        & 45.2        & 32.2        & 36.0        & 37.4        & 30.3        & \multicolumn{1}{c|}{21.0} & 34.0             \\
        \multicolumn{2}{l|}{NLLB-\med-FT (ours, used in C)}                                                                                                                & -                   & $\cmark$           & \multicolumn{1}{c|}{0.6}                                                                         & 34.4        & 38.8        & 44.6        & 34.7        & 39.0        & 41.6        & 32.1        & \multicolumn{1}{c|}{22.4} & 35.9             \\
        \multicolumn{2}{l|}{NLLB-\lrg (original, used in A, B)}                                                                                                                      & -                   & $\xmark$           & \multicolumn{1}{c|}{1.3}                                                                         & 34.6        & 38.6        & 46.8        & 33.7        & 38.2        & 39.6        & 31.8        & \multicolumn{1}{c|}{23.2} & 35.8             \\
        \multicolumn{2}{l|}{NLLB-\lrg-FT (ours, used in C)}                                                                                                                & -                   & $\cmark$           & \multicolumn{1}{c|}{1.3}                                                                         & 35.3        & 39.9        & 45.8        & 36.0        & 40.6        & 43.1        & 32.6        & \multicolumn{1}{c|}{23.9} & 37.2             \\ \midrule \midrule
        \multicolumn{14}{c}{\textit{(Other works) Zero-shot End-to-End Speech Translation}}                                                                                                                                                                                                                                                                                                                                                                            \\
        \multicolumn{2}{l|}{MultiSLT \cite{language_specific_speech}}                                                                                                   & $\cmark$            & $\cmark$           & \multicolumn{1}{c|}{0.05}                                                                        & 6.8         & 6.8         & 10.9        & -           & -           & -           & -           & \multicolumn{1}{c|}{-}    & -                \\
        \multicolumn{2}{l|}{T-Modules \cite{t_modules}}                                                                                                                 & $\cmark$            & $\xmark$           & \multicolumn{1}{c|}{2}                                                                           & 23.8        & 27.4        & 32.7        & -           & -           & -           & -           & \multicolumn{1}{c|}{-}    & -                \\
        \multicolumn{2}{l|}{DCMA \cite{dcma}}                                                                                                                           & $\cmark$            & $\xmark$           & \multicolumn{1}{c|}{0.15}                                                                        & 22.4        & 24.6        & 29.7        & 18.4        & 22.0        & 24.2        & 16.8        & \multicolumn{1}{c|}{11.8} & 21.2             \\
        \multicolumn{2}{l|}{DCMA \cite{dcma}}                                                                                                                           & $\cmark$            & $\cmark$           & \multicolumn{1}{c|}{0.15}                                                                        & 24.0        & 26.2        & 33.1        & 24.1        & 28.3        & 29.2        & 22.2        & \multicolumn{1}{c|}{16.0} & 25.4             \\
        \multicolumn{2}{l|}{SimZeroCR \cite{simreg}}                                                                                                                    & $\cmark$            & $\cmark$           & \multicolumn{1}{c|}{0.15}                                                                        & 25.1        & 27.0        & 34.6        & -           & -           & -           & -           & \multicolumn{1}{c|}{15.6} & -                \\
            \multicolumn{2}{l|}{Towards-Zero-shot \cite{towards-zero}}                                      & $\cmark$            & $\xmark$           & \multicolumn{1}{c|}{$0.1$}                                                            & 23.4                                  & 26.5                                  & 33.7                                  & -                                     & -                                     & -                                     & -                                     & \multicolumn{1}{c|}{-}                                 & -        \\
            \multicolumn{2}{l|}{Towards-Zero-shot \cite{towards-zero}}                                      & $\cmark$            & $\cmark$           & \multicolumn{1}{c|}{$0.1$}                                                            & 26.5                                  & 29.5                                  & 35.3                                  & -                                     & -                                     & -                                     & -                                     & \multicolumn{1}{c|}{-}                                 & -      
            \\
        \midrule \midrule
        \multicolumn{14}{c}{\textit{(Ours) Zero-shot End-to-End Speech Translation}}                                                                                                                                                                                                                                                                                                                                                                                   \\
        \begin{tabular}[c]{@{}l@{}}\textsc{\phantom{ii}ZeroSwot-Small}\\ \phantom{ii}(w2v-\textsc{Base} and NLLB-\med)\end{tabular}          & \multicolumn{1}{r|}{(B)} & $\cmark$            & $\xmark$           & \multicolumn{1}{c|}{0.1 / 0.7}                                                                   & 27.0        & 31.6        & 35.6        & 26.6        & 30.9        & 31.5        & 25.1        & \multicolumn{1}{c|}{17.9} & 28.3             \\ \midrule
        \multirow{3}{*}{\begin{tabular}[c]{@{}l@{}}\textsc{\phantom{ii}ZeroSwot-Medium}\\ \phantom{ii}(w2v-\lrg and NLLB-\med)\end{tabular}} & \multicolumn{1}{r|}{(A)} & $\xmark$            & $\xmark$           & \multicolumn{1}{c|}{0.35 / 0.95}                                                                 & 24.8        & 30.0        & 32.6        & 24.1        & 28.6        & 28.8        & 22.9        & \multicolumn{1}{c|}{16.4} & 26.0             \\
                                                                                                                                             & \multicolumn{1}{r|}{(B)} & $\cmark$            & $\xmark$           & \multicolumn{1}{c|}{0.35 / 0.95}                                                                 & 28.5        & 33.1        & 37.5        & 28.2        & 32.3        & 32.9        & 26.0        & \multicolumn{1}{c|}{18.7} & 29.6             \\
                                                                                                                                             & \multicolumn{1}{r|}{(C)} & $\cmark$            & $\cmark$           & \multicolumn{1}{c|}{0.35 / 0.95}                                                                 & 30.5        & 34.9        & 39.4        & 30.6        & 35.0        & 37.1        & 27.8        & \multicolumn{1}{c|}{20.3} & 31.9             \\ \midrule
        \multirow{3}{*}{\begin{tabular}[c]{@{}l@{}}\textsc{\phantom{ii}ZeroSwot-\lrg}\\ \phantom{ii}(w2v-\lrg and NLLB-\lrg)\end{tabular}}   & \multicolumn{1}{r|}{(A)} & $\xmark$            & $\xmark$           & \multicolumn{1}{c|}{0.35 / 1.65}                                                                 & 26.5        & 31.1        & 33.5        & 25.4        & 29.9        & 30.6        & 24.3        & \multicolumn{1}{c|}{18.0} & 27.4             \\
                                                                                                                                             & \multicolumn{1}{r|}{(B)} & $\cmark$            & $\xmark$           & \multicolumn{1}{c|}{0.35 / 1.65}                                                                 & 30.1        & 34.8        & 38.9        & 29.8        & 34.4        & 35.3        & 27.6        & \multicolumn{1}{c|}{20.4} & 31.4             \\
                                                                                                                                             & \multicolumn{1}{r|}{(C)} & $\cmark$            & $\cmark$           & \multicolumn{1}{c|}{0.35 / 1.65}                                                                 & 31.2        & 35.8        & 40.5        & 31.4        & 36.3        & 38.3        & 28.0        & \multicolumn{1}{c|}{21.5} & 32.9             \\ \midrule \midrule
        \multicolumn{14}{c}{\textit{(Ours) Cascaded Speech Translation with CTC ASR model}}                                                                                                                                                                                                                                                                                                                                                                            \\
        \multirow{3}{*}{\phantom{ii}\wv and NLLB-\med}                                                                                       & \multicolumn{1}{r|}{(A)} & $\xmark$            & $\xmark$           & \multicolumn{1}{c|}{(0.3 + 0.6) / 0.9}                                                           & 21.9        & 24.4        & 27.5        & 19.3        & 23.7        & 24.0        & 19.8        & \multicolumn{1}{c|}{14.2} & 21.8             \\
                                                                                                                                             & \multicolumn{1}{r|}{(B)} & $\cmark$            & $\xmark$           & \multicolumn{1}{c|}{(0.3 + 0.6) / 0.9}                                                           & 26.1        & 28.1        & 32.6        & 22.9        & 27.9        & 27.9        & 23.5        & \multicolumn{1}{c|}{16.6} & 25.7             \\
                                                                                                                                             & \multicolumn{1}{r|}{(C)} & $\cmark$            & $\cmark$           & \multicolumn{1}{c|}{(0.3 + 0.6) / 0.9}                                                           & 25.1        & 26.8        & 30.1        & 20.9        & 27.6        & 28.5        & 20.7        & \multicolumn{1}{c|}{17.8} & 24.7             \\ \midrule
        \multirow{3}{*}{\phantom{ii}\wv and NLLB-\lrg}                                                                                       & \multicolumn{1}{r|}{(A)} & $\xmark$            & $\xmark$           & \multicolumn{1}{c|}{(0.3 + 1.3) / 1.6}                                                           & 24.7        & 27.1        & 30.3        & 21.6        & 26.8        & 27.4        & 22.9        & \multicolumn{1}{c|}{16.9} & 24.7             \\
                                                                                                                                             & \multicolumn{1}{r|}{(B)} & $\cmark$            & $\xmark$           & \multicolumn{1}{c|}{(0.3 + 1.3) / 1.6}                                                           & 27.8        & 29.9        & 34.4        & 24.6        & 29.6        & 30.0        & 25.6        & \multicolumn{1}{c|}{18.5} & 27.5             \\
                                                                                                                                             & \multicolumn{1}{r|}{(C)} & $\cmark$            & $\cmark$           & \multicolumn{1}{c|}{(0.3 + 1.3) / 1.6}                                                           & 27.0        & 28.4        & 31.9        & 22.6        & 29.5        & 31.4        & 22.1        & \multicolumn{1}{c|}{19.0} & 26.5             \\ \midrule
        \multicolumn{14}{c}{\textit{(Ours) Cascaded Speech Translation with Attention-based Encoder-Decoder ASR model}}                                                                                                                                                                                                                                                                                                                                                           \\
        \multirow{3}{*}{\begin{tabular}[c]{@{}l@{}}\phantom{ii}w2v-Transformer \\ \phantom{ii}and NLLB-\med\end{tabular}}                    & \multicolumn{1}{r|}{(A)} & $\xmark$            & $\xmark$           & \multicolumn{1}{c|}{(0.4 + 0.6) / 1}                                                             & 24.3        & 27.9        & 33.2        & 23.0        & 27.0        & 27.2        & 22.5        & \multicolumn{1}{c|}{16.7} & 25.2             \\
                                                                                                                                             & \multicolumn{1}{r|}{(B)} & $\cmark$            & $\xmark$           & \multicolumn{1}{c|}{(0.4 + 0.6) / 1}                                                             & 29.0        & 32.6        & 39.8        & 27.9        & 32.0        & 32.9        & 27.1        & \multicolumn{1}{c|}{18.9} & 30.0             \\
                                                                                                                                             & \multicolumn{1}{r|}{(C)} & $\cmark$            & $\cmark$           & \multicolumn{1}{c|}{(0.4 + 0.6) / 1}                                                             & 31.1        & 34.4        & 41.4        & 30.2        & 34.7        & 37.2        & 28.3        & \multicolumn{1}{c|}{20.6} & 32.2             \\ \midrule
        \multirow{3}{*}{\begin{tabular}[c]{@{}l@{}}\phantom{ii}w2v-Transformer\\ \phantom{ii}and NLLB-\lrg\end{tabular}}                     & \multicolumn{1}{r|}{(A)} & $\xmark$            & $\xmark$           & \multicolumn{1}{c|}{(0.4 + 1.3) / 1.7}                                                           & 25.8        & 29.0        & 34.5        & 24.4        & 28.7        & 27.5        & 23.6        & \multicolumn{1}{c|}{18.2} & 26.5             \\
                                                                                                                                             & \multicolumn{1}{r|}{(B)} & $\cmark$            & $\xmark$           & \multicolumn{1}{c|}{(0.4 + 1.3) / 1.7}                                                           & 30.8        & 33.5        & 41.6        & 29.4        & 33.2        & 33.3        & 28.1        & \multicolumn{1}{c|}{20.2} & 31.3             \\
                                                                                                                                             & \multicolumn{1}{r|}{(C)} & $\cmark$            & $\cmark$           & \multicolumn{1}{c|}{(0.4 + 1.3) / 1.7}                                                           & 32.0        & 35.3        & 42.5        & 31.2        & 36.0        & 38.4        & 28.8        & \multicolumn{1}{c|}{21.6} & 33.2             \\ \midrule \midrule
        \multicolumn{14}{c}{\textit{(Other works) Supervised End-to-End Speech Translation}}                                                                                                                                                                                                                                                                                                                                                                           \\
        \multicolumn{2}{l|}{Chimera \cite{chimera}}                                                                                                                     & \multicolumn{2}{c}{$\cmark$}             & \multicolumn{1}{c|}{0.15}                                                                        & 27.1        & 30.6        & 35.6        & 25.0        & 29.2        & 30.2        & 24.0        & \multicolumn{1}{c|}{17.4} & 27.4             \\
        \multicolumn{2}{l|}{STEMM \cite{stemm}}                                                                                                                         & \multicolumn{2}{c}{$\cmark$}             & \multicolumn{1}{c|}{0.15}                                                                        & 28.7        & 31.0        & 37.4        & 25.8        & 30.5        & 31.7        & 24.5        & \multicolumn{1}{c|}{17.8} & 28.4             \\
        \multicolumn{2}{l|}{ConST \cite{const}}                                                                                                                         & \multicolumn{2}{c}{$\cmark$}             & \multicolumn{1}{c|}{0.15}                                                                        & 28.3        & 32.0        & 38.3        & 27.2        & 31.7        & 33.1        & 25.6        & \multicolumn{1}{c|}{18.9} & 29.4             \\
        \multicolumn{2}{l|}{STPT \cite{stpt}}                                                                                                                           & \multicolumn{2}{c}{$\cmark$}             & \multicolumn{1}{c|}{0.15}                                                                        & 29.2        & 33.1        & 39.7        & -           & -           & -           & -           & \multicolumn{1}{c|}{-}    & -                \\
        \multicolumn{2}{l|}{SpeechUT \cite{speechut}}                                                                                                                   & \multicolumn{2}{c}{$\cmark$}             & \multicolumn{1}{c|}{0.15}                                                                        & 30.1        & 33.6        & 41.4        & -           & -           & -           & -           & \multicolumn{1}{c|}{-}    & -                \\
        \multicolumn{2}{l|}{CMOT \cite{cmot}}                                                                                                                           & \multicolumn{2}{c}{$\cmark$}             & \multicolumn{1}{c|}{0.15}                                                                        & 29.0        & 32.8        & 39.5        & 27.5        & 32.1        & 33.5        & 26.0        & \multicolumn{1}{c|}{19.2} & 30.0             \\
        \multicolumn{2}{l|}{Siamese-PT \cite{ctc-ot}}                                                                                                                   & \multicolumn{2}{c}{$\cmark$}             & \multicolumn{1}{c|}{0.25}                                                                        & 27.9        & 31.8        & 39.2        & 27.7        & 31.7        & 34.2        & 27.0        & \multicolumn{1}{c|}{18.5} & 29.8             \\
        \multicolumn{2}{l|}{CRESS \cite{cress}}                                                                                                                         & \multicolumn{2}{c}{$\cmark$}             & \multicolumn{1}{c|}{0.15}                                                                        & 29.4        & 33.2        & 40.1        & 27.6        & 32.2        & 33.6        & 26.4        & \multicolumn{1}{c|}{19.7} & 30.3             \\
        \multicolumn{2}{l|}{SimRegCR \cite{simreg}}                                                                                                                     & \multicolumn{2}{c}{$\cmark$}             & \multicolumn{1}{c|}{0.15}                                                                        & 29.2        & 33.0        & 40.0        & 28.2        & 32.7        & 34.2        & 26.7        & \multicolumn{1}{c|}{20.1} & 30.5             \\
        \multicolumn{2}{l|}{LST (LLaMA2-7B) \cite{lst}}                                                                                                                 & \multicolumn{2}{c}{$\cmark$}             & \multicolumn{1}{c|}{7 / 7.3}                                                                     & 29.2        & 33.1        & 40.8        & -           & -           & -           & -           & \multicolumn{1}{c|}{-}    & -                \\
        \multicolumn{2}{l|}{LST (LLaMA2-13B) \cite{lst}}                                                                                                                & \multicolumn{2}{c}{$\cmark$}             & \multicolumn{1}{c|}{13 / 13.3}                                                                   & 30.4        & 35.3        & 41.6        & -           & -           & -           & -           & \multicolumn{1}{c|}{-}    & -                \\ \midrule \midrule
        \multicolumn{14}{c}{\textit{(Ours) Experiments on Compression with \zeroswot-\med (B) from Table \ref{tab:compression_results}}}                                                                                                                                                                                                                                                                                                                               \\
        \multicolumn{2}{l|}{Without Compression}                                                                                                                        & $\cmark$            & $\xmark$           & \multicolumn{1}{c|}{0.35 / 0.95}                                                                 & 27.6        & 33.0        & 36.4        & 27.4        & 31.5        & 32.0        & 25.2        & \multicolumn{1}{c|}{18.2} & 28.9             \\
        \multicolumn{2}{l|}{Length Adaptor (down-sampling $\times 4$)}                                                                                                  & $\cmark$            & $\xmark$           & \multicolumn{1}{c|}{0.35 / 0.95}                                                                 & 27.8        & 32.4        & 36.7        & 27.9        & 31.4        & 31.6        & 25.4        & \multicolumn{1}{c|}{18.3} & 28.9             \\
        \multicolumn{2}{l|}{Character-level Compression}                                                                                                                & $\cmark$            & $\xmark$           & \multicolumn{1}{c|}{0.35 / 0.95}                                                                 & 27.9        & 32.6        & 36.9        & 27.8        & 31.8        & 31.4        & 25.4        & \multicolumn{1}{c|}{18.1} & 29.0             \\
        \multicolumn{2}{l|}{Compr. Adapter + Word Tokenization}                                                                                                            & $\cmark$            & $\xmark$           & \multicolumn{1}{c|}{0.35 / 0.95}                                                                 & 26.4        & 31.0        & 33.7        & 25.7        & 29.6        & 30.3        & 24.1        & \multicolumn{1}{c|}{17.3} & 27.2             \\
        \multicolumn{2}{l|}{Compr. Adapter + Subword Tokenization}                                                                                                  & $\cmark$            & $\xmark$           & \multicolumn{1}{c|}{0.35 / 0.95}                                                                 & 27.3        & 32.5        & 35.4        & 26.4        & 31.0        & 31.6        & 25.0        & \multicolumn{1}{c|}{18.3} & 28.4             \\ \midrule
        \multicolumn{14}{c}{\textit{(Ours) Ablations with \zeroswot-\med (B) from Table \ref{tab:ablations}}}                                                                                                                                                                                                                                                                                                                                           \\
        \multicolumn{2}{l|}{Without Speech Embedder}                                                                                                                    & $\cmark$            & $\xmark$           & \multicolumn{1}{c|}{0.35 / 0.95}                                                                 & 27.2        & 32.6        & 36.0        & 27.1        & 31.3        & 31.9        & 25.1        & \multicolumn{1}{c|}{17.9} & 28.6             \\
        \multicolumn{2}{l|}{Without Auxiliary Wasserstein Losses}                                                                                                       & $\cmark$            & $\xmark$           & \multicolumn{1}{c|}{0.35 / 0.95}                                                                 & 28.4        & 33.1        & 37.3        & 27.9        & 32.3        & 32.6        & 25.9        & \multicolumn{1}{c|}{18.6} & 29.5             \\ \bottomrule \bottomrule
        \end{tabular}
        }
        \caption{Extended Results on \mustc{} \tstcommon.}
        \label{tab:extended_results_mustc}
    \end{table*}

    \begin{table*}[ht]
        \centering
        \resizebox{0.9\textwidth}{!}
        {
        \begin{tabular}{@{}ccccccccccccc@{}}
        \toprule
        \multirow{2}{*}{\textbf{Models}} & \multirow{2}{*}{\textbf{Type}} & \multirow{2}{*}{\textbf{ASR Data}} & \multirow{2}{*}{\textbf{\begin{tabular}[c]{@{}c@{}}Training\\ Params (B)\end{tabular}}} & \multicolumn{9}{c}{\textbf{WER}}                                                                                                               \\ \cmidrule(l){5-13} 
                                         &                                &                                    &                                                                                         & \textbf{De} & \textbf{Es} & \textbf{Fr} & \textbf{It} & \textbf{Nl} & \textbf{Pt} & \textbf{Ro} & \textbf{Ru}               & \textbf{Average} \\ \midrule
        w2v                              & CTC                            & \cv                                & \multicolumn{1}{c|}{0.3}                                                                & 42.5        & 42.5        & 42.5        & 42.6        & 42.6        & 42.6        & 42.5        & \multicolumn{1}{c|}{42.5} & 42.5             \\
        w2v                              & CTC                            & \textsc{MuST-C}                             & \multicolumn{1}{c|}{0.3}                                                                & 21.4        & 21.4        & 21.4        & 21.4        & 21.4        & 21.5        & 21.5        & \multicolumn{1}{c|}{21.5} & 21.4             \\
        w2v-Transformer                  & AED                            & \cv                                & \multicolumn{1}{c|}{0.4}                                                                & 20.4        & 20.9        & 20.5        & 20.7        & 20.3        & 21.1        & 20.8        & \multicolumn{1}{c|}{20.6} & 20.7             \\
        w2v-Transformer                  & AED                            & \textsc{MuST-C}                             & \multicolumn{1}{c|}{0.4}                                                                & 9.8         & 10.1        & 9.8         & 10.0        & 9.6         & 10.6        & 10.2        & \multicolumn{1}{c|}{9.9}  & 10.0             \\ \bottomrule
        \end{tabular}
        }
        \caption{ASR Results on \mustc{} \tstcommon. AED stands for attention-based encoder-decoder.}
        \label{tab:asr_results_mustc}
    \end{table*}

\section{Detailed Results in \covost} \label{appendix:results_covost_full}

    Here we provide a more extended version of Table \ref{tab:main_results_covost}, including MT results of NLLB (original \& finetuned), \zeroswot results with the original NLLB, and also add XLS-R-0.3B \cite{xls-r}. 

    \begin{table*}[ht]
        \centering
        \resizebox{\textwidth}{!}
        {
        \begin{tabular}{@{}lccccccccccccccccc@{}}
        \toprule
        \multicolumn{1}{c}{\textbf{Models}}      & \textbf{Size (B)} & \textbf{Ar} & \textbf{Ca} & \textbf{Cy} & \textbf{De} & \textbf{Et} & \textbf{Fa} & \textbf{Id} & \textbf{Ja} & \textbf{Lv} & \multicolumn{1}{l}{\textbf{Mn}} & \textbf{Sl} & \textbf{Sv} & \textbf{Ta} & \textbf{Tr} & \textbf{Zh} & \textbf{Average} \\ \midrule
        \multicolumn{18}{c}{\textit{Machine Translation}}                                                                                                                                                                                                                                                                     \\
        NLLB-M (original)                        & 0.6               & 20.0        & 39.0        & 26.3        & 35.5        & 23.4        & 15.7        & 39.6        & 21.8        & 14.8        & 10.4                            & 30.3        & 41.1        & 20.2        & 21.1        & 34.8        & 26.3             \\
        NLLB-M-\textsc{CoVoST} (ours)            & 0.6               & 28.5        & 46.3        & 35.5        & 37.1        & 31.5        & 29.2        & 45.2        & 38.4        & 29.1        & 22.0                            & 37.7        & 45.4        & 29.9        & 23.0        & 46.7        & 35.0             \\
        NLLB-L (original)                        & 1.3               & 23.3        & 43.5        & 33.5        & 37.9        & 27.9        & 16.6        & 41.9        & 23.0        & 20.0        & 13.1                            & 35.1        & 43.8        & 21.7        & 23.8        & 37.5        & 29.5             \\
        NLLB-L-\textsc{CoVoST} (ours)            & 1.3               & 29.9        & 47.8        & 35.6        & 38.8        & 32.7        & 29.9        & 46.4        & 39.5        & 29.9        & 21.7                            & 39.3        & 46.8        & 31.0        & 24.4        & 48.2        & 36.1             \\ \midrule
        \multicolumn{18}{c}{(Ours) Zero-shot End-to-End Speech Translation}                                                                                                                                                                                                                                                   \\
        \textsc{ZeroSwot-M}                      & 0.35 / 0.95       & 17.6        & 32.5        & 18.0        & 29.9        & 20.4        & 16.3        & 32.4        & 32.0        & 13.3        & 10.0                            & 25.2        & 34.4        & 17.8        & 15.6        & 30.5        & 23.1             \\
        $\hookrightarrow$ NLLB-M-\textsc{CoVoST} & 0.35 / 0.95       & 24.4        & 38.7        & 28.8        & 31.2        & 26.2        & 26.0        & 36.0        & 46.0        & 24.8        & 19.0                            & 31.6        & 37.8        & 24.4        & 18.6        & 39.0        & 30.2             \\
        \textsc{ZeroSwot-L}                      & 0.35 / 1.65       & 19.8        & 36.1        & 22.6        & 31.8        & 23.6        & 16.8        & 34.2        & 33.6        & 17.5        & 11.8                            & 28.9        & 36.8        & 19.1        & 17.5        & 32.2        & 25.5             \\
        $\hookrightarrow$ NLLB-L-\textsc{CoVoST} & 0.35 / 1.65       & 25.7        & 40.0        & 29.0        & 32.8        & 27.2        & 26.6        & 37.1        & 47.1        & 25.7        & 18.9                            & 33.2        & 39.3        & 25.3        & 19.8        & 40.5        & 31.2             \\ \midrule
        \multicolumn{18}{c}{(Other works) Supervised End-to-End Speech Translation}                                                                                                                                                                                                                                           \\
        XLS-R-0.3B                               & 0.3               & 16.3        & 28.7        & 29.1        & 23.6        & 19.6        & 19.0        & 27.4        & 36.9        & 19.3        & 13.2                            & 22.4        & 29.1        & 15.6        & 15.0        & 33.5        & 23.2             \\
        XLS-R-1B                                 & 1.0               & 19.2        & 32.1        & 31.8        & 26.2        & 22.4        & 21.3        & 30.3        & 39.9        & 22.0        & 14.9                            & 25.4        & 32.3        & 18.1        & 17.1        & 36.7        & 26.0             \\
        XLS-R-2B                                 & 2.0               & 20.7        & 34.2        & 33.8        & 28.3        & 24.1        & 22.9        & 32.5        & 41.5        & 23.5        & 16.2                            & 27.6        & 34.5        & 19.8        & 18.6        & 38.5        & 27.8             \\
        \textsc{SeamlessM4T-M}                   & 1.2               & 20.8        & 37.3        & 29.9        & 31.4        & 23.3        & 17.2        & 34.8        & 37.5        & 19.5        & 12.9                            & 29.0        & 37.3        & 18.9        & 19.8        & 30.0        & 26.6             \\
        \textsc{SeamlessM4T-L}                   & 2.3               & 24.5        & 41.6        & 33.6        & 35.9        & 28.5        & 19.3        & 39.0        & 39.4        & 23.8        & 15.7                            & 35.0        & 42.5        & 22.7        & 23.9        & 33.1        & 30.6             \\
        $\hookrightarrow$ v2                     & 2.3               & 25.4        & 43.6        & 35.5        & 37.0        & 29.3        & 19.2        & 40.2        & 39.7        & 24.8        & 16.4                            & 36.2        & 43.7        & 23.4        & 24.7        & 35.9        & 31.7             \\ \bottomrule
        \end{tabular}
        }
        \caption{Extended Results on \covost{} \test.}
        \label{tab:extended_results_covost}
    \end{table*}

\section{Detailed Results in FLEURS} \label{appendix:results_fleurs_full}

    In Table \ref{tab:extended_results_fleurs} we present the per-language\footnote{\href{https://github.com/openlanguagedata/flores}{github.com/openlanguagedata/flores}} results for FLEURS En-X test. \zeroswot is trained on \cv, while "+ More Data" indicates that it was additionally trained on \mustcasr and \ls. NLLB results are obtained by us, by running the original versions, while \seamless \cite{seamlessm4t} results are obtained from their repository\footnote{\href{https://github.com/facebookresearch/seamless_communication/blob/main/docs/m4t/README.md}{facebookresearch/seamless\_communication}}.

\begin{table*}[ht]
        \centering
        \resizebox{\textwidth}{!}
        {
\begin{tabular}{@{}l|ccccc|ccccc@{}}
\toprule
\multirow{2}{*}{\textbf{\begin{tabular}[c]{@{}l@{}}Language\\ Code\end{tabular}}} & \multicolumn{5}{c|}{\med}                                                                          & \multicolumn{5}{c}{\lrg}                                                                           \\ \cmidrule(l){2-11} 
                                                                                  & \textbf{NLLB} & \textbf{Cascade} & \textbf{\zeroswot} & \textbf{+ More Data} & \textbf{\seamless} & \textbf{NLLB} & \textbf{Cascade} & \textbf{\zeroswot} & \textbf{+ More Data} & \textbf{\seamless} \\ \midrule
amh-Ethi                                                                          & 12.2          & 9.2              & 9.9               & 10.0                 & 10.0                 & 13.2          & 11.5             & 9.4               & 9.4                  & 12.2                 \\
arb-Arab                                                                          & 23.1          & 18.1             & 18.7              & 18.9                 & 20.0                 & 26.4          & 20.5             & 21.2              & 21.9                 & 22.9                 \\
asm-Beng                                                                          & 6.8           & 4.9              & 6.6               & 6.7                  & 6.0                  & 6.8           & 6.2              & 7.1               & 6.7                  & 6.7                  \\
azj-Latn                                                                          & 11.0          & 8.3              & 8.7               & 8.6                  & 9.2                  & 13.2          & 10.2             & 10.4              & 10.4                 & 11.0                 \\
bel-Cyrl                                                                          & 11.3          & 8.7              & 8.3               & 8.5                  & 8.6                  & 13.3          & 10.2             & 9.5               & 9.9                  & 10.3                 \\
ben-Beng                                                                          & 14.8          & 12.0             & 13.5              & 13.7                 & 13.2                 & 17.6          & 13.6             & 15.0              & 15.2                 & 15.1                 \\
bos-Latn                                                                          & 26.9          & 20.8             & 20.9              & 21.7                 & 24.3                 & 31.2          & 23.2             & 24.5              & 25.4                 & 26.8                 \\
bul-Cyrl                                                                          & 36.7          & 29.4             & 29.8              & 30.7                 & 30.4                 & 40.4          & 33.4             & 33.2              & 33.5                 & 35.5                 \\
cat-Latn                                                                          & 37.1          & 28.6             & 30.6              & 30.7                 & 32.7                 & 40.4          & 32.0             & 33.6              & 34.2                 & 35.8                 \\
ceb-Latn                                                                          & 28.6          & 22.2             & 21.6              & 22.2                 & 21.6                 & 29.6          & 25.0             & 22.2              & 22.3                 & 22.2                 \\
ces-Latn                                                                          & 28.4          & 21.1             & 20.7              & 21.7                 & 22.6                 & 30.2          & 24.6             & 23.3              & 24.4                 & 24.8                 \\
ckb-Arab                                                                          & 8.3           & 5.7              & 7.5               & 7.7                  & 8.2                  & 10.8          & 7.7              & 8.3               & 8.5                  & 10.0                 \\
zho-Hans                                                                          & 28.2          & 27.1             & 24.4              & 24.9                 & 25.8                 & 31.7          & 30.2             & 26.1              & 25.4                 & 29.8                 \\
cym-Latn                                                                          & 33.2          & 25.3             & 26.2              & 26.7                 & 33.7                 & 41.5          & 32.4             & 33.0              & 33.4                 & 37.4                 \\
dan-Latn                                                                          & 40.3          & 32.8             & 33.9              & 34.4                 & 34.7                 & 42.5          & 35.9             & 35.6              & 36.1                 & 39.2                 \\
deu-Latn                                                                          & 34.8          & 28.1             & 27.7              & 28.7                 & 28.6                 & 37.4          & 31.2             & 31.0              & 31.4                 & 32.5                 \\
ell-Grek                                                                          & 24.0          & 20.0             & 20.4              & 20.7                 & 19.4                 & 26.5          & 22.2             & 22.5              & 23.0                 & 22.0                 \\
est-Latn                                                                          & 18.4          & 14.2             & 14.1              & 14.7                 & 17.6                 & 23.4          & 18.9             & 18.5              & 18.4                 & 21.0                 \\
fin-Latn                                                                          & 18.3          & 13.9             & 14.0              & 14.4                 & 15.8                 & 22.6          & 18.2             & 18.5              & 19.0                 & 20.3                 \\
fra-Latn                                                                          & 46.2          & 37.7             & 35.5              & 36.3                 & 37.4                 & 48.7          & 40.4             & 37.4              & 38.3                 & 41.4                 \\
fuv-Latn                                                                          & 1.2           & 0.3              & 0.5               & 0.6                  & 0.7                  & 2.3           & 0.4              & 1.0               & 1.1                  & 0.7                  \\
gaz-Latn                                                                          & 3.4           & 1.9              & 1.7               & 1.7                  & 2.8                  & 3.9           & 3.2              & 2.0               & 2.0                  & 4.9                  \\
gle-Latn                                                                          & 22.5          & 16.6             & 17.0              & 17.7                 & 20.7                 & 27.4          & 21.0             & 21.1              & 21.3                 & 23.4                 \\
glg-Latn                                                                          & 30.9          & 24.7             & 25.3              & 25.8                 & 27.1                 & 33.5          & 26.5             & 27.4              & 28.1                 & 29.0                 \\
guj-Gujr                                                                          & 22.1          & 16.9             & 17.6              & 17.8                 & 17.7                 & 23.4          & 18.1             & 18.4              & 18.4                 & 19.7                 \\
heb-Hebr                                                                          & 24.2          & 18.9             & 19.3              & 19.9                 & 19.6                 & 29.3          & 23.4             & 24.3              & 24.2                 & 24.9                 \\
hin-Deva                                                                          & 29.7          & 24.7             & 24.6              & 25.3                 & 27.1                 & 30.8          & 26.5             & 25.8              & 26.0                 & 28.8                 \\
hrv-Latn                                                                          & 23.7          & 18.3             & 18.9              & 19.0                 & 21.0                 & 28.8          & 21.2             & 22.8              & 23.2                 & 22.8                 \\
hun-Latn                                                                          & 21.1          & 15.8             & 16.4              & 16.5                 & 16.7                 & 24.4          & 19.7             & 19.3              & 19.8                 & 19.7                 \\
hye-Armn                                                                          & 15.9          & 13.1             & 14.7              & 15.2                 & 14.8                 & 17.9          & 15.1             & 17.1              & 17.1                 & 16.2                 \\
ibo-Latn                                                                          & 15.6          & 13.1             & 11.6              & 11.7                 & 13.6                 & 16.7          & 14.5             & 12.0              & 11.9                 & 13.8                 \\
ind-Latn                                                                          & 42.3          & 30.8             & 31.9              & 32.4                 & 33.7                 & 44.7          & 33.2             & 34.2              & 34.4                 & 36.8                 \\
isl-Latn                                                                          & 19.5          & 14.1             & 14.7              & 15.2                 & 15.6                 & 22.3          & 17.7             & 16.8              & 17.0                 & 20.6                 \\
ita-Latn                                                                          & 28.0          & 21.5             & 22.1              & 22.4                 & 21.9                 & 29.1          & 22.6             & 23.2              & 23.7                 & 23.9                 \\
jav-Latn                                                                          & 25.5          & 16.9             & 17.8              & 17.8                 & 19.8                 & 28.0          & 19.1             & 19.4              & 19.6                 & 20.4                 \\
jpn-Jpan                                                                          & 34.1          & 31.6             & 30.5              & 30.8                 & 31.9                 & 36.2          & 34.3             & 32.9              & 33.3                 & 35.6                 \\
kan-Knda                                                                          & 17.7          & 13.1             & 13.4              & 13.6                 & 13.5                 & 19.4          & 14.3             & 14.3              & 14.2                 & 15.2                 \\
kat-Geor                                                                          & 13.0          & 10.4             & 9.9               & 9.8                  & 9.8                  & 14.0          & 11.1             & 10.6              & 10.7                 & 12.1                 \\
kaz-Cyrl                                                                          & 18.8          & 14.6             & 14.3              & 14.5                 & 15.5                 & 20.9          & 17.3             & 15.8              & 15.4                 & 18.8                 \\
khk-Cyrl                                                                          & 9.1           & 6.9              & 7.5               & 7.7                  & 9.8                  & 11.4          & 9.4              & 9.5               & 9.7                  & 12.2                 \\
khm-Khmr                                                                          & 0.8           & 0.1              & 2.5               & 2.7                  & 0.7                  & 1.4           & 0.4              & 2.6               & 2.8                  & 0.4                  \\
kir-Cyrl                                                                          & 12.3          & 9.0              & 9.2               & 9.6                  & 9.8                  & 13.0          & 10.5             & 10.1              & 10.2                 & 11.4                 \\
kor-Hang                                                                          & 12.0          & 9.2              & 9.8               & 10.1                 & 10.4                 & 12.2          & 11.5             & 10.8              & 10.7                 & 11.7                 \\
lao-Laoo                                                                          & 54.3          & 49.4             & 53.1              & 53.0                 & 54.5                 & 56.1          & 53.3             & 54.7              & 54.8                 & 55.2                 \\
lit-Latn                                                                          & 19.0          & 14.1             & 14.0              & 14.2                 & 16.1                 & 22.1          & 18.1             & 16.8              & 16.9                 & 19.0                 \\
lug-Latn                                                                          & 6.6           & 3.6              & 3.6               & 3.4                  & 6.4                  & 8.6           & 5.2              & 4.9               & 5.0                  & 6.8                  \\
luo-Latn                                                                          & 10.8          & 7.2              & 7.4               & 7.5                  & 9.6                  & 11.2          & 8.4              & 8.2               & 8.1                  & 9.2                  \\
lvs-Latn                                                                          & 18.1          & 14.5             & 15.2              & 15.5                 & 20.1                 & 23.2          & 19.7             & 19.8              & 19.9                 & 23.0                 \\
mal-Mlym                                                                          & 14.2          & 10.4             & 9.8               & 9.8                  & 11.0                 & 14.7          & 13.0             & 10.7              & 10.7                 & 13.3                 \\
mar-Deva                                                                          & 13.6          & 10.2             & 10.8              & 10.7                 & 11.4                 & 16.1          & 11.8             & 12.0              & 12.4                 & 13.0                 \\
mkd-Cyrl                                                                          & 28.9          & 23.5             & 23.7              & 23.5                 & 26.5                 & 33.3          & 27.4             & 26.9              & 27.2                 & 28.8                 \\
mlt-Latn                                                                          & 24.2          & 20.9             & 14.8              & 14.6                 & 24.0                 & 28.8          & 24.0             & 18.0              & 18.0                 & 27.6                 \\
mya-Mymr                                                                          & 41.4          & 43.1             & 40.4              & 40.3                 & 41.5                 & 43.3          & 45.8             & 43.1              & 43.3                 & 43.0                 \\
nld-Latn                                                                          & 25.3          & 19.9             & 20.8              & 21.2                 & 21.7                 & 26.6          & 20.7             & 22.6              & 22.8                 & 24.2                 \\
nob-Latn                                                                          & 30.8          & 24.7             & 25.0              & 25.6                 & 26.8                 & 33.2          & 26.4             & 26.3              & 27.1                 & 28.5                 \\
npi-Deva                                                                          & 16.7          & 12.7             & 13.2              & 13.3                 & 15.3                 & 16.9          & 14.1             & 13.1              & 13.5                 & 15.7                 \\
nya-Latn                                                                          & 14.3          & 9.6              & 9.2               & 9.1                  & 10.4                 & 14.2          & 10.9             & 9.6               & 10.1                 & 10.8                 \\
ory-Orya                                                                          & 14.2          & 11.3             & 12.3              & 12.4                 & 12.6                 & 15.6          & 13.5             & 12.5              & 12.5                 & 13.5                 \\
pan-Guru                                                                          & 22.6          & 17.7             & 18.8              & 19.5                 & 19.5                 & 24.4          & 19.0             & 21.1              & 21.1                 & 21.7                 \\
pbt-Arab                                                                          & 12.7          & 10.5             & 10.9              & 11.0                 & 11.1                 & 13.8          & 10.4             & 11.4              & 11.5                 & 11.6                 \\
pes-Arab                                                                          & 21.7          & 17.4             & 22.5              & 22.9                 & 18.3                 & 23.2          & 19.6             & 23.5              & 24.3                 & 21.1                 \\
pol-Latn                                                                          & 18.3          & 13.6             & 13.6              & 14.1                 & 14.6                 & 20.4          & 15.9             & 15.9              & 16.2                 & 16.9                 \\
por-Latn                                                                          & 45.5          & 37.0             & 37.5              & 38.2                 & 38.0                 & 47.0          & 39.2             & 39.4              & 40.0                 & 40.5                 \\
ron-Latn                                                                          & 33.5          & 28.0             & 26.1              & 26.4                 & 29.3                 & 35.6          & 32.0             & 27.8              & 28.4                 & 31.9                 \\
rus-Cyrl                                                                          & 27.5          & 21.7             & 22.1              & 22.3                 & 22.1                 & 29.8          & 24.2             & 24.6              & 25.2                 & 25.8                 \\
slk-Latn                                                                          & 27.8          & 21.0             & 21.5              & 21.7                 & 22.6                 & 32.3          & 25.1             & 25.4              & 25.7                 & 27.1                 \\

\end{tabular}
        }
    \end{table*}

\clearpage \newpage

\begin{table*}[t]
        \centering
        \resizebox{\textwidth}{!}
        {
\begin{tabular}{@{}l|ccccc|ccccc@{}}
                                                                                  & \phantom{\textbf{NLLB}} & \phantom{\textbf{Cascade}} & \phantom{\textbf{ZeroSwot}} & \phantom{\textbf{+ More Data}} & \phantom{\textbf{SeamlessM4T}} & \phantom{\textbf{NLLB}} & \phantom{\textbf{Cascade}} & \phantom{\textbf{ZeroSwot}} & \phantom{\textbf{+ More Data}} & \phantom{\textbf{SeamlessM4T}} \\
slv-Latn                                                                          & 23.3          & 17.6             & 18.1              & 18.7                 & 19.4                 & 26.0          & 20.6             & 21.6              & 22.0                 & 23.2                 \\
sna-Latn                                                                          & 11.4          & 6.7              & 6.4               & 6.8                  & 5.8                  & 11.8          & 6.9              & 6.8               & 6.4                  & 6.7                  \\
snd-Arab                                                                          & 19.4          & 16.3             & 17.1              & 17.5                 & 18.9                 & 21.0          & 17.1             & 18.3              & 18.0                 & 18.0                 \\
som-Latn                                                                          & 11.2          & 8.4              & 8.0               & 8.0                  & 8.5                  & 12.1          & 8.8              & 8.7               & 8.6                  & 9.7                  \\
spa-Latn                                                                          & 27.5          & 21.7             & 22.1              & 22.9                 & 21.1                 & 27.7          & 22.4             & 22.9              & 23.2                 & 23.9                 \\
srp-Cyrl                                                                          & 27.6          & 22.2             & 21.9              & 22.8                 & 26.8                 & 32.6          & 26.2             & 26.3              & 27.2                 & 30.3                 \\
swe-Latn                                                                          & 40.7          & 32.0             & 33.2              & 34.1                 & 34.0                 & 44.0          & 35.2             & 36.2              & 36.6                 & 38.4                 \\
swh-Latn                                                                          & 31.6          & 24.8             & 23.6              & 23.8                 & 26.2                 & 32.7          & 27.0             & 24.8              & 24.7                 & 27.9                 \\
tam-Taml                                                                          & 16.2          & 11.8             & 11.8              & 12.4                 & 13.3                 & 18.6          & 14.0             & 13.2              & 13.9                 & 15.4                 \\
tel-Telu                                                                          & 21.1          & 15.2             & 15.3              & 15.3                 & 17.4                 & 24.1          & 17.1             & 17.4              & 17.3                 & 19.4                 \\
tgk-Cyrl                                                                          & 18.6          & 14.2             & 14.9              & 15.1                 & 0.1                  & 22.0          & 17.3             & 17.2              & 17.7                 & 0.1                  \\
tgl-Latn                                                                          & 30.5          & 22.7             & 23.5              & 23.7                 & 26.1                 & 33.7          & 24.8             & 26.0              & 26.4                 & 28.0                 \\
tha-Thai                                                                          & 47.7          & 46.7             & 46.5              & 46.5                 & 49.8                 & 51.1          & 50.0             & 49.5              & 49.9                 & 51.2                 \\
tur-Latn                                                                          & 23.1          & 16.5             & 17.5              & 18.0                 & 19.4                 & 27.4          & 20.4             & 21.1              & 21.4                 & 22.6                 \\
ukr-Cyrl                                                                          & 23.0          & 18.5             & 19.0              & 18.8                 & 20.8                 & 26.9          & 22.0             & 22.3              & 22.2                 & 24.8                 \\
urd-Arab                                                                          & 21.3          & 18.0             & 18.4              & 18.7                 & 18.9                 & 23.2          & 19.6             & 20.4              & 20.6                 & 20.6                 \\
uzn-Latn                                                                          & 14.3          & 11.3             & 11.3              & 11.0                 & 12.8                 & 17.2          & 13.5             & 13.6              & 13.6                 & 14.6                 \\
vie-Latn                                                                          & 38.1          & 31.3             & 32.2              & 32.6                 & 32.7                 & 40.8          & 33.6             & 34.4              & 35.0                 & 35.1                 \\
yor-Latn                                                                          & 3.7           & 3.1              & 3.2               & 3.0                  & 4.4                  & 5.0           & 2.1              & 4.2               & 3.8                  & 4.5                  \\
zho-Hant                                                                          & 0.5           & 0.1              & 1.2               & 1.2                  & 0.2                  & 0.6           & 0.2              & 1.2               & 1.2                  & 0.3                  \\
zsm-Latn                                                                          & 37.2          & 28.1             & 28.1              & 28.7                 & 21.5                 & 40.6          & 30.7             & 30.2              & 30.4                 & 34.2                 \\
zul-Latn                                                                          & 16.2          & 11.1             & 11.0              & 11.4                 & 11.9                 & 18.7          & 11.4             & 12.2              & 12.3                 & 13.4                 \\ \midrule
Average                                                                           & 22.5          & 17.8             & 18.1              & 18.4                 & 19.2                 & 24.8          & 20.2             & 20.1              & 20.3                 & 21.5                 \\ \bottomrule
\end{tabular}
        }
        \caption{Extended Results on FLEURS \test.}
        \label{tab:extended_results_fleurs}
    \end{table*}


\end{document}